\documentclass[sigconf]{acmart}

\usepackage{algorithm}
\usepackage{algorithmic}
\usepackage{makecell}
\usepackage{array}
\usepackage{graphics}
\usepackage{subcaption}
\usepackage{multirow}
\usepackage{float}
\usepackage{enumitem}
\usepackage{tikz}
\usepackage{balance}
\usepackage{placeins}
\usepackage{pifont}

\definecolor{IncrGreen}{rgb}{0.25, 0.55, 0.25}
\AtBeginDocument{
  }

\copyrightyear{2026}
\acmYear{2026}
\setcopyright{cc}
\setcctype{by}
\acmConference[KDD '26]{Proceedings of the 32nd ACM SIGKDD Conference on Knowledge Discovery and Data Mining V.2}{August 09--13, 2026}{Jeju Island, Republic of Korea}
\acmBooktitle{Proceedings of the 32nd ACM SIGKDD Conference on Knowledge Discovery and Data Mining V.2 (KDD '26), August 09--13, 2026, Jeju Island, Republic of Korea}
\acmDOI{10.1145/3770855.3818450}
\acmISBN{979-8-4007-2259-2/2026/08}

\begin{document}

\title[MLT-Dedup: Efficient Large-Scale Online Video Deduplication]{MLT-Dedup: Efficient Large-Scale Online Video Deduplication via Multi-Level Representations and Spatial-Temporal Matching}

\author{David Yuchen Wang}
\authornote{Equal contribution.}
\affiliation{
  \institution{TikTok}
  \city{Singapore}
  \country{Singapore}
}
\email{david.w@tiktok.com}

\author{Haoying Li}
\authornotemark[1]
\affiliation{
  \institution{TikTok}
  \city{Singapore}
  \country{Singapore}
}
\email{haoying.li@tiktok.com}

\author{Hailun Xu}
\affiliation{
  \institution{TikTok}
  \city{Singapore}
  \country{Singapore}
}
\email{xuhailun@tiktok.com}

\author{Wei Chee Yew}
\affiliation{
  \institution{TikTok}
  \city{Singapore}
  \country{Singapore}
}
\email{weichee.yew@tiktok.com}

\author{Zirui Zhu}
\affiliation{
  \institution{TikTok}
  \city{Singapore}
  \country{Singapore}
}
\affiliation{
  \department{School of Computing}
  \institution{National University of Singapore}
  \city{Singapore}
  \country{Singapore}
}
\email{ziruizhu@u.nus.edu}

\author{Sanjay Saha}
\affiliation{
  \institution{TikTok}
  \city{Singapore}
  \country{Singapore}
}
\email{sanjay.saha@tiktok.com}

\author{Hao Hei}
\affiliation{
  \institution{TikTok}
  \city{Singapore}
  \country{Singapore}
}
\email{hao.hei1@tiktok.com}

\author{Kanchan Sarkar}
\authornote{Project lead.}
\authornote{Corresponding author.}
\affiliation{
  \institution{TikTok}
  \city{San Jose}
  \state{CA}
  \country{USA}
}
\email{kanchan.sarkar@tiktok.com}

\author{Kun Xu}
\affiliation{
  \institution{TikTok}
  \city{San Jose}
  \state{CA}
  \country{USA}
}
\email{daniel.chen28@tiktok.com}

\renewcommand{\shortauthors}{David Yuchen Wang et al.}

\begin{abstract}
The explosive growth of user-generated video content on online platforms is accompanied by the emergence of numerous near-duplicate videos—videos that are identical or highly similar but differ by partial edits. These duplicates degrade user experience and increase storage and bandwidth costs, making large-scale video deduplication a critical task. Existing video deduplication frameworks face a fundamental challenge in retrieving sufficient high-quality candidates under a limited index budget, as well as trade-offs between efficiency and precision. To address these issues, we propose \textbf{MLT-Dedup}, an efficient large-scale online video deduplication framework with \textbf{M}ulti-\textbf{L}evel representations and spatial-\textbf{T}emporal matching. Our approach employs a \textbf{M}ulti-\textbf{L}evel \textbf{V}ideo \textbf{E}ncoder (\textbf{ML-VE}) to extract both fine-grained frame-level and sparse clip-level embeddings: sparse embeddings support efficient candidate retrieval, while fine-grained embeddings are loaded for precise pairwise matching. During matching, we introduce \textbf{DiF-SiM}, a \textbf{Di}fferential \textbf{F}eature-enhanced \textbf{Si}milarity \textbf{M}odule capable of locating duplicated temporal segments and providing reliable similarity evidence to support policy-driven deduplication decisions. Extensive experiments on a real-world large-scale platform demonstrate that MLT-Dedup reduces online repetition rates by \textbf{91\%} at 90\% precision. Furthermore, our sparse retrieval design achieves a \textbf{5$\times$} increase in indexing capacity, enabling broader candidate coverage in real-world deployment.

\end{abstract}



\maketitle

\section{Introduction}
With the proliferation of multi-source user-generated content, a large number of near-duplicate videos emerge, differing only by lightweight transformations such as clipping, watermark insertion, or format conversion \cite{rodrigues2010equal,liu2015quantitative}. These duplicated videos degrade search and recommendation quality while increasing copyright risks and storage overhead, making video deduplication critical at scale. In practice, video deduplication frameworks consist of three key components: video representation, candidate retrieval, and video pair matching \cite{black2023vader, kordopatis2022dns, shen2020advance}. Specifically, videos are encoded into visual representations, indexed for scalable retrieval, and further examined by matching models to make deduplication decisions. In large-scale user-generated-content (UGC) platforms, deduplication systems must operate under strict index resources while meeting strong requirements on duplicate reduction, as inefficient designs directly reduce candidate coverage and deduplication accuracy. In this paper, we propose an efficient large-scale online video deduplication framework, MLT-Dedup, to address the indexing capacity issue while reducing video repetition via multi-granularity representation and spatial--temporal matching. 
\begin{figure*}[t]
  \centering
  \includegraphics[width=\linewidth,trim={1cm 0.5cm 0.5cm 0.2cm},clip]{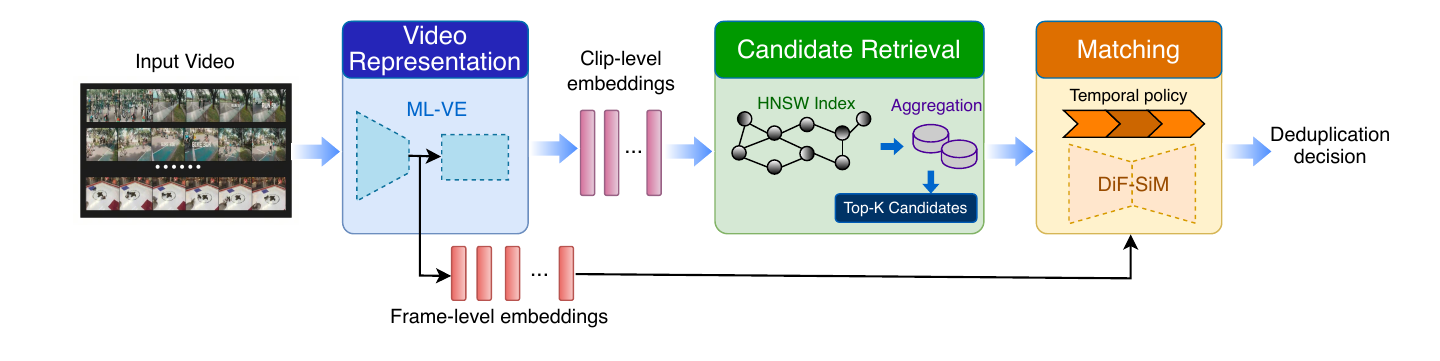}
  \caption{Overall framework of MLT-Dedup: Input videos are encoded by ML-VE into fine-grained frame-level embeddings (used for precise pairwise matching) and sparse clip-level embeddings (indexed via HNSW for high-recall candidate retrieval). Retrieved top-K candidates are matched using DiF-SiM, which computes temporal overlaps to make deduplication decisions.}
  \label{fig:framework}
\end{figure*}

In the video representation stage of deduplication frameworks, visual embeddings are typically generated by a recall model and can take two forms: frame-level and clip-level embeddings \cite{kordopatis2019finding,kordopatis2022dns,black2023vader,shen2020advance,lin2025mmembeduniversalmultimodalretrieval}. Frame-level embeddings extract features for every frame, providing fine-grained visual details that benefit accurate duplicate detection. Yet under a fixed indexing budget, such dense representations severely limit index capacity, resulting in fewer retrievable candidates and degraded recall. Clip-level embeddings summarize video segments through key-frame selection \cite{zhu2026focus} or direct clip encoding. While clip-level embeddings significantly reduce memory footprint and indexing cost, their coarse granularity may discard fine temporal details, potentially degrading matching accuracy. This trade-off reveals a key challenge in large-scale deduplication frameworks. To address this challenge, we deploy a multi-level video encoder, ML-VE, which can generate frame-level and clip-level embeddings simultaneously. Coarse clip-level embeddings are used for scalable indexing and retrieval, while fine-grained frame-level embeddings are stored on disk and only loaded for matching when a candidate is retrieved. To ensure effective retrieval performance, ML-VE is trained with dedicated optimization strategies, including hybrid loss formulations, inner-feature knowledge distillation, and data-centric memory bank deduplication.

In the video pair matching stage, video duplicate detection methods relying on global similarity \cite{near_duplicate_video_retreival_with_deep_metric_learning,bardes2022vicreg, practical_elimination_of_near_duplicates_from_web_video_search} degrade as video editing techniques become more sophisticated. They cannot provide temporal localization of duplicated content and are prone to misclassifying partially overlapping videos due to lack of fine-grained temporal information and sensitivity to irrelevant content. To tackle this issue, we introduce the proportion of overlapping content policy to the video pair matching stage, that is, only candidates exceeding a certain temporal coverage are considered duplicates. This not only ensures that partially overlapping videos are not misclassified, but is also fair when comparing videos with varying lengths and supports policy-driven content management such as manual review. To enforce such policies effectively, we propose DiF-SiM, a differential feature-enhanced similarity module in the video pair matching stage. DiF-SiM captures temporal relations between video frames through differential features ($\Delta \mathbf{f}_t$), a learned similarity metric, and pre-training techniques leveraging unlabeled videos and images.

To summarize, this paper makes the following contributions:
\vspace{-0.3cm}
\begin{itemize}
    \item We propose MLT-Dedup, which addresses the indexing capacity issue while reducing video repetition by integrating compact multi-level video representation and a carefully designed spatial--temporal matching model.

    \item In the video representation stage, we introduce ML-VE along with effective training strategies to generate video embeddings at both coarse and fine granularities. Coarse embeddings are used for retrieval to improve candidate coverage and recall, while fine-grained embeddings enhance matching precision in downstream deduplication.

    \item In the pairwise matching stage, we propose DiF-SiM, capable of locating duplicated temporal segments between video pairs and providing reliable similarity evidence to support policy-driven deduplication decisions.
\end{itemize}

Extensive experiments on a large-scale online platform show that MLT-Dedup reduces the online repetition rate by 91\% at 90\% precision, while achieving a 5$\times$ increase in indexing capacity under the same resource budget.

\section{Method}
\subsection{Overall Framework}
MLT-Dedup follows the standard three-stage video deduplication pipeline, consisting of video representation, candidate retrieval, and pairwise matching, as shown in Fig. \ref{fig:framework}. Given an input video, ML-VE first encodes a video as multi-level representations. Compact clip-level embeddings are indexed using a Hierarchical Navigable Small World (HNSW) graph-based approximate nearest neighbor index \cite{hnsw} to retrieve top-K candidate embeddings, which are subsequently aggregated to obtain candidate videos. Fine-grained frame-level embeddings are stored on disk and loaded on demand for precise matching. Retrieved candidate video embeddings are then examined by a fine-grained matching module, DiF-SiM, which localizes duplicated temporal overlap and outputs a similarity score. Finally, policy-driven deduplication decisions are made based on the proportion of overlapping time. The following subsections describe the design of the ML-VE model, the clip-level embedding-based retrieval strategy, and the DiF-SiM matching module in detail.

\subsection{Multi-Level Video Representations}
The challenge of building effective large-scale video retrieval systems lies in maintaining high retrieval recall under a fixed resource budget. With limited memory and storage, the library can only store a small number of representations per video, which reduces visual coverage and candidate recall. To address this trade-off, we propose ML-VE, a video encoder that generates multi-level video representations in a single forward pass. Compact clip-level embeddings are used for scalable retrieval, while fine-grained frame-level embeddings are retained for precise downstream matching. The effectiveness of these multi-level representations is achieved through a jointly designed model architecture and a set of targeted training strategies, such as hybrid loss formulation, inner-feature knowledge distillation, and data-centric memory bank deduplication.
\begin{figure}[t]
  \centering
  \includegraphics[width=\linewidth,trim={1cm 0cm 1.2cm 0.5cm},clip]{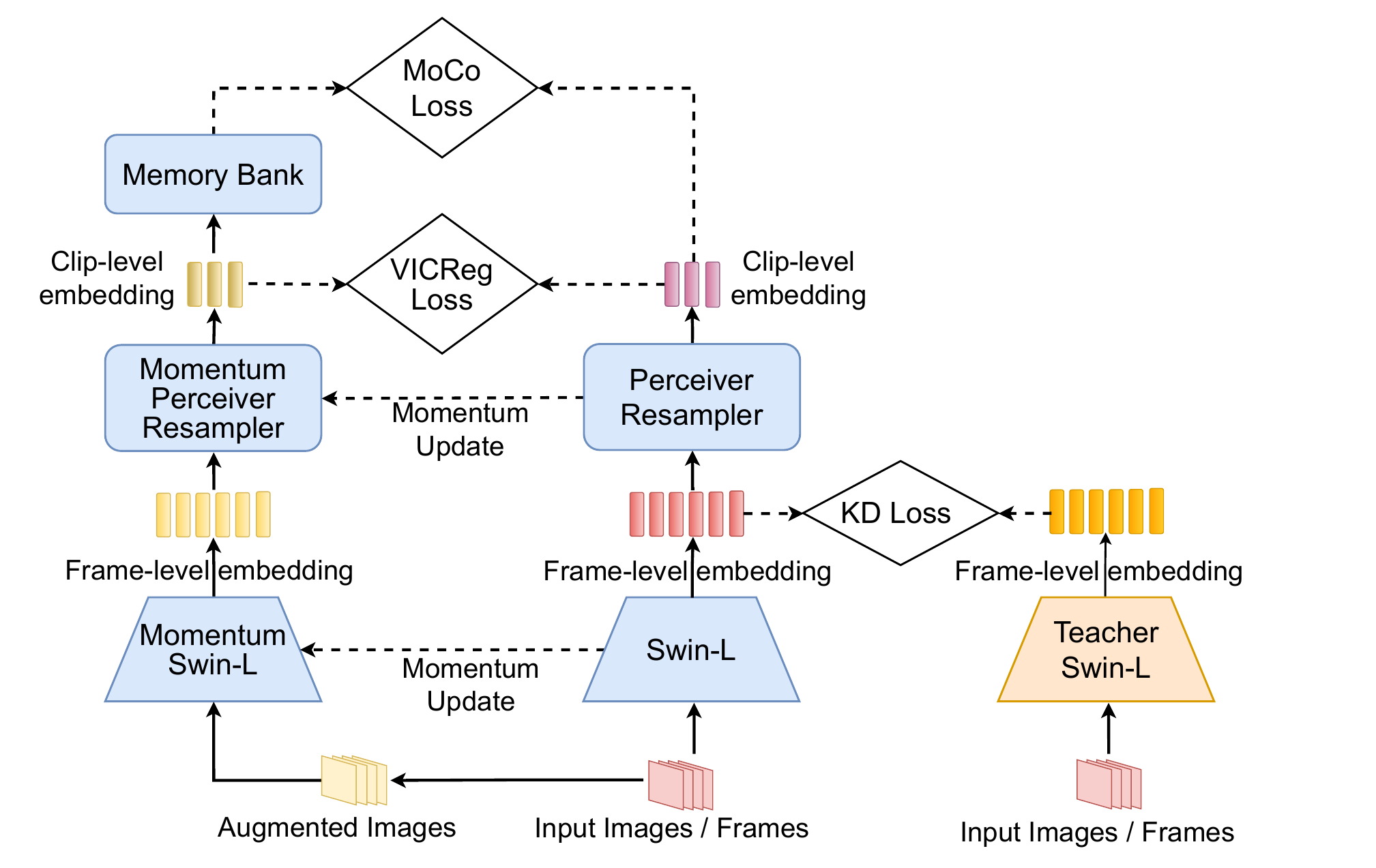}
  \caption{ML-VE architecture.}
  \Description{A model architecture diagram illustrating how ML-VE generates multi-level video representations for retrieval and matching.}
  \label{fig:ME-recall}
\end{figure}
\vspace{-0.1cm}
\subsubsection{ML-VE Architecture}
ML-VE is built upon a Swin-Large backbone~\cite{liu2021swintransformerhierarchicalvision} and trained under a MoCo-style momentum contrastive learning framework~\cite{he2020momentumcontrastunsupervisedvisual}, as illustrated in Figure~\ref{fig:ME-recall}.
For each video frame, the \emph{online} (query) branch encodes the input with a Swin-Large encoder $B_{\theta}$, producing frame-level visual tokens, and then applies a Perceiver Resampler $R_{\theta}$ to aggregate temporally distributed features into a fixed number of latent tokens.
This yields compact clip-level token representations that are subsequently average-pooled into a $D$-dim global embedding for large-scale indexing and candidate retrieval.
In parallel, we maintain a \emph{momentum} (key) branch with the same architecture, parameterized by $\bar{\theta}$, which is updated as an exponential moving average (EMA) of the online parameters.
During training, the momentum branch produces stable key embeddings that are enqueued into a memory bank (FIFO queue) to construct a large set of negative samples for contrastive learning. Our basic training objective consists of a VICReg \cite{bardes2022vicreg} regularization loss and a MoCo \cite{he2020momentumcontrastunsupervisedvisual} contrastive loss.
\vspace{-0.2cm}
\paragraph{\textbf{VICReg loss:}}
We apply VICReg on paired embeddings from the two branches:
{\small
\begin{equation}
\mathcal{L}_{\mathrm{VICReg}}
=\lambda\,\mathcal{L}_{\mathrm{inv}}
+\mu\,\mathcal{L}_{\mathrm{var}}
+\nu\,\mathcal{L}_{\mathrm{cov}},
\end{equation}}
where $\lambda,\mu,\nu$ are scalar weights. Let $Q\in\mathbb{R}^{B\ding{55} D}$ and $K\in\mathbb{R}^{B\ding{55} D}$ be the batch-stacked
embeddings whose $b$-th rows are $q_b^\top$ and $k_b^\top$, respectively. We define
{\small
\setlength{\abovedisplayskip}{3pt}
\setlength{\belowdisplayskip}{3pt}
\setlength{\abovedisplayshortskip}{2pt}
\setlength{\belowdisplayshortskip}{2pt}
\begin{align}
\mathcal{L}_{\mathrm{inv}} &= \frac{1}{B}\sum_{b=1}^{B}\|q_b-k_b\|_2^2, \\
\mathcal{L}_{\mathrm{var}} &=
\frac{1}{D}\sum_{d=1}^{D}\Big[\max\big(0,\gamma-\sigma(Q_{:,d})\big)+\max\big(0,\gamma-\sigma(K_{:,d})\big)\Big], \\
\mathcal{L}_{\mathrm{cov}} &=
\frac{1}{D}\sum_{i\neq j}\left(\mathrm{Cov}(Q)_{ij}^2+\mathrm{Cov}(K)_{ij}^2\right),
\end{align}
}
where $\sigma(Q_{:,d})$ denotes the standard deviation of the $d$-th feature over the minibatch,
$\gamma>0$ is a margin, and $\mathrm{Cov}(Q),\mathrm{Cov}(K)\in\mathbb{R}^{D\ding{55} D}$ are covariance matrices computed on centered
embeddings:
{\small
\begin{equation}
\mathrm{Cov}(Q)=\frac{1}{B-1}\tilde{Q}^{\top}\tilde{Q},\qquad
\mathrm{Cov}(K)=\frac{1}{B-1}\tilde{K}^{\top}\tilde{K},
\end{equation}}
with $\tilde{Q}$ and $\tilde{K}$ obtained by subtracting the per-dimension batch mean from $Q$ and $K$.
\vspace{-0.1cm}
\paragraph{\textbf{MoCo loss:}}
We maintain a memory bank (queue) $\mathcal{Q}$ of size $K$ that stores negative keys $k^{-}\in\mathbb{R}^{D}$ produced by the
momentum branch. With temperature $\tau$ and similarity function $\mathrm{sim}(\cdot,\cdot)$, the MoCo
(InfoNCE) loss is defined as
\begin{equation}
\mathcal{L}_{\mathrm{MoCo}}
=
-\frac{1}{B}\sum_{b=1}^{B}
\log
\frac{\exp(\mathrm{sim}(q_b,k_b)/\tau)}
{\exp(\mathrm{sim}(q_b,k_b)/\tau)+\sum\limits_{k^{-}\in\mathcal{Q}}\exp(\mathrm{sim}(q_b,k^{-})/\tau)}.
\end{equation}
\vspace{-0.5cm}
\subsubsection{Hybrid Loss Formulation}
To further enhance the discriminative power of the learned embeddings for retrieval and matching, we incorporate a supervised pairwise loss and a triplet loss on top of the self-supervised pre-training. This formulation combines the generalization ability of self-supervised representation learning with task-specific supervision, encouraging the learned embeddings to better distinguish near-duplicate and non-duplicate videos in retrieval and matching scenarios.
\vspace{-0.1cm}
\paragraph{\textbf{Pairwise Loss:}}
We use a labeled pair dataset $\mathcal{D}_{\mathrm{pair}}=\{(v_i,v_j,y_{ij})\}$ with $y_{ij}\in\{0,1\}$ indicating whether $(v_i,v_j)$ is a near-duplicate pair.
Let $q_i,q_j\in\mathbb{R}^{D}$ be the pooled embeddings produced by the online branch for $v_i$ and $v_j$.
Using similarity $s_{ij}=\mathrm{sim}(q_i,q_j)$ (consistent with MoCo), we optimize a logistic pairwise loss:
{\small
\begin{equation}
\mathcal{L}_{\mathrm{match}}
=
-\frac{1}{|\mathcal{D}_{\mathrm{pair}}|}
\sum_{(i,j)}
\Big[
y_{ij}\log \sigma(s_{ij}) + (1-y_{ij})\log\!\big(1-\sigma(s_{ij})\big)
\Big],
\end{equation}}
where $\sigma(\cdot)$ is the sigmoid function.
\vspace{-0.1cm}
\paragraph{\textbf{Triplet Loss:}}
We form triplets $\mathcal{D}_{\mathrm{trip}}=\{(v_a,v_p,v_n)\}$ from labels, where $(v_a,v_p)$ is a positive (duplicate) pair and $(v_a,v_n)$ is a negative pair.
Let $q_a,q_p,q_n\in\mathbb{R}^{D}$ denote their online pooled embeddings. We use a margin-based triplet loss:
{\small
\begin{equation}
\mathcal{L}_{\mathrm{tri}}
=
\frac{1}{|\mathcal{D}_{\mathrm{trip}}|}
\sum_{(a,p,n)}
\max\Big(0,\ \mathrm{sim}(q_a,q_n)-\mathrm{sim}(q_a,q_p)+m\Big),
\end{equation}}
where $m$ is the margin.
This objective enforces that positive pairs are embedded closer than negative ones by a fixed margin.
\vspace{-0.1cm}
\subsubsection{Inner-Feature Knowledge Distillation}
In practice, we observe that joint optimization with retrieval-oriented objectives biases ML-VE toward clip-level representations, while the quality of frame-level embeddings degrades, leading to suboptimal downstream matching performance. To explicitly enhance frame-level features, we introduce an inner-feature knowledge distillation (KD) strategy.

We adopt a Swin-Large teacher model trained to produce matching-friendly frame-level embeddings. The teacher model is trained on 7M image samples using a MoCo contrastive loss applied to the average-pooled token embedding, encouraging strong local visual semantics without clip-level aggregation bias.
During ML-VE training, we apply feature-level distillation at multiple intermediate layers of the Swin-Large backbone.
Let $f^{(l)}_{\theta}$ and $f^{(l)}_{\mathrm{T}}$ denote the student and teacher feature maps at layer $l$.
The distillation loss is defined as:
{\small
\begin{equation}
\mathcal{L}_{\mathrm{KD}}
=
\sum_{l \in \mathcal{L}}
\left\|
\mathrm{Norm}\big(f^{(l)}_{\theta}\big)
-
\mathrm{Norm}\big(f^{(l)}_{\mathrm{T}}\big)
\right\|_2^2.
\end{equation}
}

This inner-feature supervision improves the discriminability of frame-level embeddings for matching, without increasing indexing memory or affecting retrieval efficiency.

To summarize, the overall objective is:
{\small
\begin{equation}
\mathcal{L}
=
\mathcal{L}_{\mathrm{MoCo}}
+\alpha\,\mathcal{L}_{\mathrm{VICReg}}
+\beta\,\mathcal{L}_{\mathrm{match}}
+\rho\,\mathcal{L}_{\mathrm{tri}}
+\omega\,\mathcal{L}_{\mathrm{KD}},
\end{equation}}
where $\alpha$, $\beta$, $\rho$, $\omega$ balance each loss term.

\subsubsection{Data-Centric Memory Bank Deduplication}
As commonly observed in large-scale contrastive learning, random sampling over massive datasets inevitably introduces near-duplicated samples, reducing the number of unique negatives and limiting contrastive supervision.
To mitigate this issue, we adopt a query-adaptive filtering strategy when computing the contrastive loss with the memory bank.
For each query embedding, we compute cosine similarities against all key embeddings in the memory bank. Based on the similarity scores, we identify near-positive or highly similar samples and exclude them from the negative set. The contrastive loss is then computed only over the remaining dissimilar keys, which serve as valid negatives for the given query.
This data-centric, query-level filtering preserves the diversity of the memory bank while preventing false negatives from dominating the training signal.

\subsection{Sparse Retrieval}
\label{HNSW_Retrieval_and_Clip-to-Video_Aggregation}
We perform retrieval on sparse clip-level representations from ML-VE, rather than dense frame-level embeddings, to improve indexing efficiency and candidate recall coverage under a fixed memory budget. 
Specifically, compact clip-level embeddings are indexed using an HNSW-based approximate nearest neighbor structure \cite{faiss_gpu}, enabling scalable retrieval with low memory overhead while allowing a larger portion of the video corpus to be retained in the index. Given a query video, its clip-level embeddings are independently queried against the HNSW index to retrieve nearest-neighbor embeddings. Each retrieved embedding corresponds to a specific clip from a candidate video. We then aggregate these clip-level matches into video-level similarity scores through a clip-to-video aggregation process. Retrieved results are first grouped by candidate video ID. If multiple clips from the same candidate video are retrieved by the same query clip, only the match with the highest similarity score is retained to avoid redundant contributions from temporally adjacent segments. Subsequently, for each candidate video, we collect its retained similarity scores across all query clips, discard low-confidence matches below a similarity threshold (i.e., 0.4), and average the remaining scores to obtain a single video-level similarity score. Finally, candidate videos are ranked by their aggregated similarity scores, and the top-$K$ videos are selected as retrieval results for downstream fine-grained matching. This coarse-to-fine design follows a similar efficiency principle to recent multi-stage content decision systems~\cite{zhu2026camel,yew2025dynamic}.

\subsection{Spatial-Temporal Matching}
In this stage, we determine duplication based on the proportion of overlapping content in time and deduplication score: \textbf{a video is flagged as a duplicate only when the duplicated temporal coverage exceeds a predefined threshold, and has a deduplication score over the threshold.}
This strategy avoids misclassifying partially overlapping videos, enables fair comparison across videos of different lengths, and provides explicit temporal localization for downstream auditing and manual review.

We observe that using clip-level embeddings from ML-VE directly for matching leads to degraded accuracy, especially for partially duplicated or temporally misaligned videos. Frame-level embeddings better preserve fine-grained visual and temporal details, enabling more accurate modeling of localized content reuse. Thus, the spatial-temporal matching stage performs fine-grained verification on frame-level representations. Only the embeddings of retrieved video pairs are loaded and processed in this stage, which allows the matching stage to remain compute-efficient.

To effectively leverage these fine-grained spatial-temporal matching signals, we propose DiF-SiM, a differential feature-enhanced similarity module for spatial-temporal video matching. DiF-SiM models frame-level correspondences between video pairs, localizes duplicated temporal segments, and produces similarity evidence that directly supports proportion-based deduplication policies.
The architecture and matching strategy are detailed as follows.

\begin{figure}[t]
\centering
\includegraphics[width=0.5\textwidth,trim={0cm 0.2cm 0.2cm 0.2cm},clip]{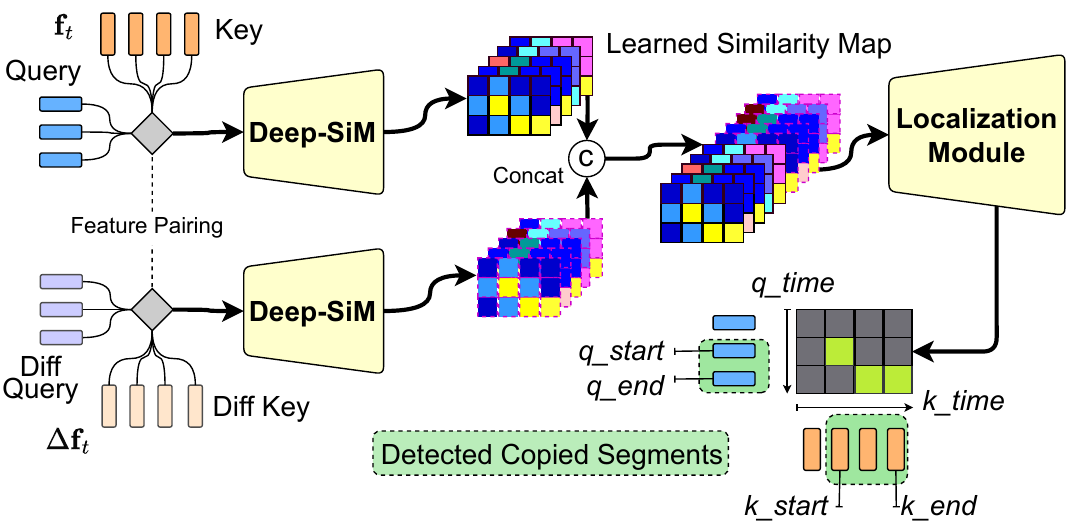}
\caption{DiF-SiM model architecture: Static frame-level features from the ML-VE are used with their differential features ($\Delta \mathbf{f}_t$) to produce similarity maps by Deep-Sim. They are then concatenated and fed into a localization module to predict the temporal boundaries of duplicated segments that dictate the timestamps q\_start, q\_end, k\_start, and k\_end.}
\Description{An architecture diagram showing static and differential frame-level features flowing through Deep-Sim and a localization module to estimate duplicated temporal boundaries.}
\label{fig:dif_sim}
\end{figure}

\subsubsection{DiF-SiM Model Architecture}
As shown in Figure \ref{fig:dif_sim}, frame-level embeddings from ML-VE are used as inputs of DiF-SiM. To model temporal correspondence, we also compute differential features by taking frame-wise differences of the embeddings, which highlight local temporal changes and are complementary to the static appearance cues. Both sets of features are passed to a deep similarity module (Deep-Sim), which learns an enhanced similarity map between query and key feature pairs. The learned similarity maps are concatenated with cosine similarity maps and passed through a localization module to detect copied temporal video segments. Finally, DiF-SiM outputs the estimated overlap duration for both videos and a deduplication score ([q\_start, q\_end, k\_start, k\_end, score]). Only when both the predicted overlap duration and the deduplication score exceed predefined thresholds could we consider the video pair to be duplicate.

\subsubsection{Differential Features}
In spatial-temporal video matching, directly comparing query and candidate videos via independent frame-to-frame similarity is insufficient for accurately modeling partial duplication and temporal misalignment.
Such static pairwise comparisons ignore temporal transitions within each video, making it difficult to distinguish true content reuse from incidental visual similarity and to robustly localize duplicated segments.

To address this limitation, we leverage both conventional static frame features and additional \emph{differential features} that capture video dynamics within each video. 
Given a sequence of frame-level embeddings $\mathbf{f}_t \in \mathbb{R}^D, t = 1, \dots, T,$ extracted by ML-VE, we compute normalized differences between consecutive frames:
\begin{small}
\begin{equation}
\Delta \mathbf{f}_t = \mathrm{Norm}(\mathbf{f}_{t+1} - \mathbf{f}_t),
\end{equation}
\end{small}
where $\mathrm{Norm}(\cdot)$ denotes $\ell_2$ normalization.
For the last frame, we apply wrap-around to keep the feature length consistent:
$\Delta \mathbf{f}_T = \mathrm{Norm}(\mathbf{f}_1 - \mathbf{f}_T)$.

The resulting differential features characterize local temporal transitions and motion patterns within a video, explicitly encode motion cues between adjacent frames, and complement static appearance information encoded in the original static embeddings. As illustrated in Appendix \ref{sec:similarity_visualization}, similarity maps derived from differential features exhibit clearer temporal alignment, especially under partial duplication and temporal shifts.

\subsubsection{Deep-Sim: Deep Similarity Module}
In the spatial-temporal matching stage, similarity maps are computed to reflect temporal alignment. However, directly computing similarity using a fixed cosine metric between feature vectors is often insufficient, because it compresses rich relationships into a single scalar value and lacks the flexibility to integrate heterogeneous cues. To handle this issue, we introduce a \emph{Deep Similarity Module} (Deep-Sim), which learns a parametric similarity metric that maps a pair of features to a high-dimensional similarity representation. Instead of relying on a hand-crafted metric, Deep-Sim models similarity as a learnable function, enabling richer semantic alignment between video elements.

\begin{figure}[h!]
\centering
\includegraphics[width=0.5\textwidth,trim={0.6cm 0.2cm 0.1cm 0.4cm},clip]{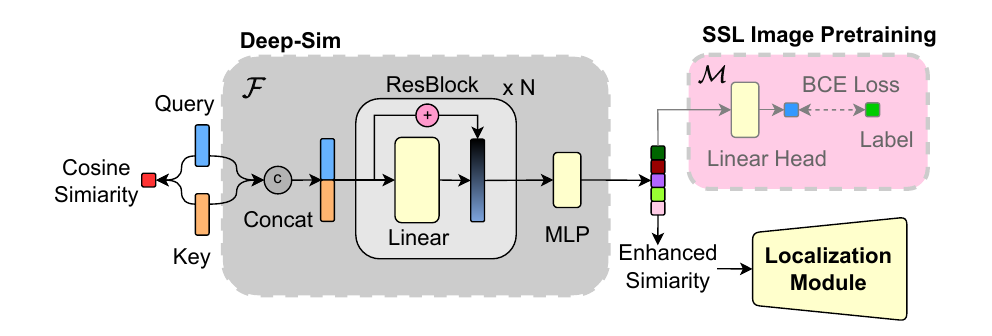}
\caption{Deep-Sim model architecture.} 
\Description{A neural network diagram showing paired feature inputs processed by residual blocks and an MLP to produce learned similarity representations.}
\label{fig:deep_sim}
\end{figure}

The architecture of the Deep-Sim module is shown in Figure \ref{fig:deep_sim}. The input query and candidate features are concatenated and passed through a series of residual linear layers (Res-Block), followed by a multi-layer perceptron (MLP). The module learns an enhanced similarity for each query-key pair, achieving strong semantics compared to direct cosine similarity. During SSL pre-training, an additional MLP is introduced to enhance model training by using transformed image pairs in place of the query and key.

We pre-train Deep-Sim in a self-supervised manner using unlabeled image data to endow it with general-purpose similarity modeling capability. Specifically, we attach an auxiliary linear head $\mathcal{M}$ to the Deep-Sim network $\mathcal{F}$. Deep-Sim is optimized by a binary cross-entropy loss to distinguish matching and non-matching pairs. This pre-training stage encourages Deep-Sim to learn discriminative similarity patterns beyond simple feature distance. After pre-training, we discard the auxiliary head $\mathcal{M}$ and transfer the learned weights of $\mathcal{F}$ into the DiF-SiM module. In the full system, Deep-Sim is fine-tuned jointly with the downstream localization module, allowing similarity estimation to adapt to video-level temporal matching objectives.

Deep-Sim is applied to both static frame features and differential features, producing two enhanced similarity maps that capture complementary static and temporal correspondences. In parallel, we compute a conventional cosine similarity map on both static frame features and differential features. These four similarity maps are concatenated along the channel dimension and fed into the localization module to infer duplicated temporal segments.

\subsubsection{Localization Module}
Our localization module follows previous works \cite{transvcl, eccv_paper} and use the YOLOX-s \cite{yolox} detector with width$=0.33$, depth$=0.33$, with an input shape of $640 \times 640$. To accommodate similarity-map inputs, we modify the shape of the first convolutional layer accordingly. The localization module performs detection on a two-dimensional time plane—each bounding-box corresponds to a "time-time" rectangular window (horizontal axis for query time, vertical axis for candidate time), indicating that two time intervals may match with each other. The localization module outputs the estimated overlap boundaries for both videos, together with a deduplication confidence score.

\subsubsection{Self-Supervised pre-training with Dual-Modality Data Augmentation}
To better initialize the model and reduce reliance on the number of paired labels, we pre-train DiF-SiM via self-supervised learning (SSL) \cite{ssl_survey, chen2020simple, ssl_first}. In raw unlabeled video collections, true duplicated pairs are extremely sparse. Manual data annotation for the video copy localization task is difficult and expensive. To address this, we leverage two modalities of unlabeled data, images and videos, to improve model performance via self-supervised learning (SSL) during model pre-training.

\paragraph{\textbf{Video SSL}}
We leverage unlabeled video data to pre-train the entire DiF-SiM under realistic copy-and-edit scenarios.
Inspired by prior work~\cite{eccv_paper}, we design three lightweight video-level augmentation strategies, as illustrated in Appendix \ref{appendix:ssl-aug}.
\begin{itemize}
    \item \textit{Self-insertion} generates duplicate pairs by randomly selecting and reinserting segments within the same video, modeling rapid edits and temporal shifts.
    \item \textit{Single-insertion} inserts segments from one video into another, combined with visual and temporal perturbations, to simulate common cross-video duplication such as reuploads with trimming or speed changes.
    \item \textit{Double-insertion} inserts segments from a source video into two different background videos, producing challenging cases with high semantic similarity but different contexts.
\end{itemize}

\vspace{-0.1cm}
\paragraph{\textbf{Image SSL}}
We leverage unlabeled image data to pre-train the Deep-Sim module. For each mini-batch of unlabeled images, we apply a random augmentation pipeline (i.e., cropping, flipping, color jittering) to each image and feed the augmented image into ML-VE to obtain an embedding. It forms a positive pair, labeled as 1, that comes from the same original image. A negative pair comes from a different image within this batch, labeled as 0.

Together, these transformations expose the model to diverse duplication patterns and strengthen its ability to localize copied regions under temporal misalignment and visual variations.

\section{Experiments}
\subsection{Datasets}
\paragraph{\textbf{ML-VE model}}
ML-VE is trained on a comprehensive dataset combining 24 million in-house videos and 4.6 billion open-source images from LAION-2B \cite{schuhmann2022laion}. Most of the in-house videos are unlabeled, and are used for self-supervised training. A small portion of labeled pairwise and triplet data are used for supervised learning. The test set includes 1k in-house query videos, each having 20 to 100 candidates.
\vspace{-0.2cm}
\paragraph{\textbf{DiF-SiM model}}
In production, DiF-SiM is trained on 0.5M human-annotated in-house video clip pairs, together with 2M image samples from Conceptual Captions~\cite{cc2m} and 3M video samples from the InternVid dataset~\cite{internvid} for self-supervised pre-training. All videos are first processed by the ML-VE model to extract 256-d frame-level embeddings, which serve as inputs to DiF-SiM. Labels are discarded for the SSL data.
\vspace{-0.2cm}
\subsection{Implementation Details}
For both ML-VE and DiF-SiM, training utilizes 16 NVIDIA A100 GPUs. We use the AdamW optimizer~\cite{adamW} with a learning rate of $0.0002$, betas of $[0.9, 0.999]$, weight decay of $0.01$, and epsilon of $1e^{-8}$. We utilize a cosine annealing learning rate scheduler with $100$ warm-up steps. For each video clip, we regard 16 frames as a clip. ML-VE outputs 768-dimensional clip-level embeddings for downstream retrieval, and 256-dimensional frame-level embeddings for DiF-SiM matching. DiF-SiM finally outputs a $640\times640$ map indicating overlap times and duplication scores.
\vspace{-0.2cm}
\subsection{Evaluation Results}
\subsubsection{Overall Online Performance}
To evaluate real-world impact, we conducted online A/B tests by incrementally deploying each component. The experiments ran for several weeks, with 10\% of traffic allocated to each experimental group. We evaluate online performance using the repetition rate, defined as the proportion of duplicated video pairs among all distinct video pairs formed from videos recommended to the same user, where duplication is determined via human annotation. Statistical significance is assessed using a two-sample z-test for proportions. The baseline system employs Swin-Large~\cite{liu2021swintransformerhierarchicalvision} to extract frame-level video representations and uses a non-parametric cosine matcher for fine-grained similarity computation by directly comparing frame embeddings. All online results are reported at a fixed precision of 90\%.

As shown in Table~\ref{tab:online_main}, MLT-Dedup system achieves substantial reductions in content repetition. Replacing the baseline representation model with ML-VE alone reduces the repetition rate by 72\%, showing that sparse clip-level retrieval dramatically improves candidate coverage and effectively mitigates missed duplicates at the retrieval stage. Replacing the cosine matcher with DiF-SiM yields a 24\% reduction in repetition rate. This result indicates that learned, parametric spatial-temporal verification can filter out a significant portion of false positives that cannot be resolved by static, frame-wise similarity measures. By jointly integrating ML-VE and DiF-SiM, MLT-Dedup achieves a total repetition rate reduction of \textbf{91\%}. The gains arise from stage-wise synergy, where ML-VE enhances candidate retrieval and DiF-SiM refines spatial-temporal matching.
In addition, the sparse clip-level representations produced by ML-VE significantly improve indexing efficiency. With a fixed amount of index resources, MLT-Dedup stores up to \textbf{5$\times$} more content in the retrieval index, enabling a longer index time-to-live (TTL) without sacrificing deduplication accuracy.

\begin{table}[t]
\centering
\small
\setlength{\tabcolsep}{6pt}
\renewcommand{\arraystretch}{1.03}
\caption{Online deduplication performance of the overall pipeline at 90\% precision. We show the reduction in repetition rates due to the addition of our ML-VE and DiF-SiM matcher to our baseline. ``Swin-L'' denotes ``Swin-Large''.}
\label{tab:online_main}
\begin{tabular}{l |c  c}
\toprule
Method & Repetition Rate & Index Size \\
\midrule
Swin-L + Cosine Matcher (Baseline) & --      & $1$ \\
ML-VE + Cosine Matcher                    & -72\%   & $\times5$ \\
Swin-L + DiF-SiM                   & -24\%   & $\times1$ \\
ML-VE + DiF-SiM (MLT-Dedup)               & \textbf{-91\%} & $\times5$ \\
\bottomrule
\end{tabular}
\end{table}
\vspace{1cm}
\begin{table}[t]
\centering
\small
\setlength{\tabcolsep}{6pt}
\renewcommand{\arraystretch}{1.03}
\caption{Comparison between different video representation models for recall evaluated on our in-house data.}
\label{tab:embed_models_recall5}
\begin{tabular}{l |c c c}
\toprule
Model & Frames/Emb & Dimension & Recall@5 \\
\midrule
Webssl-Dino-7b-Full-8b-518     & 1  & 4096 & 82.60 \\
Dinov2-Large                   & 1  & 1024 & 85.49 \\
GME-Qwen2-VL-2B                & 1  & 3584 & 83.67 \\
MM\_Embed                      & 1  & 4096 & 82.76 \\
VLM2Vec                        & 1  & 4096 & 80.68 \\
E5-V                           & 1  & 4096 & 79.91 \\
\textbf{ML-VE (ours)}    & \textbf{16} & \textbf{768}  & \textbf{87.59} \\
\bottomrule
\end{tabular}
\end{table}
\vspace{-0.3cm}
\begin{table}[h]\small
\caption{Comparison of the performance on the VCSL test set of our DiF-SiM model against open-source baselines. All methods utilize the same set of frame level embeddings extracted by the ISC encoder.}
\label{tab:vcsl-results}
\centering
\begin{tabular}{l | c >{\centering\arraybackslash}p{2cm} c}
\toprule
 Method & F-score & Precision & Recall \\
\midrule
HV  & 51.73 &  36.82 &  86.94 \\
TN  & 64.43 &  66.50 &  62.49 \\
DP  & 54.53 &  60.63 &  49.56 \\
DTW & 50.23 &  56.67 &  45.10 \\
SPD & 62.97 &  56.27 &  71.47 \\
TransVCL & 66.51 &  67.46 &  65.59  \\
RTR    & 69.22  &  66.77 &  71.85    \\ 
RTR + pre-training & 70.73  & 68.19 &  73.46 \\
\textbf{DiF-SiM (ours)}& \textbf{74.31} & \textbf{71.21} &  \textbf{77.68} \\
\bottomrule
\end{tabular}
\end{table}
\vspace{-0.8cm}
\subsubsection{Evaluation of ML-VE}
We compare our proposed model with open-source video representation models (Webssl-Dino-7b-Full-8b-518 \cite{fan2025scalinglanguagefreevisualrepresentation},Dinov2-Large \cite{oquab2024dinov2learningrobustvisual}, GME-Qwen2-VL-2B\cite{zhang2025gmeimprovinguniversalmultimodal}, MM\_Embed\cite{lin2025mmembeduniversalmultimodalretrieval}, E5-V \cite{jiang2024e5vuniversalembeddingsmultimodal})  under Recall@5, which measures the proportion of ground-truth relevant videos that appear in the top-5 retrieved results. Each method is evaluated on the same 1k clusters. As shown in Table~\ref{tab:embed_models_recall5}, despite using the coarse temporal granularity of one embedding per 16 frames, ML-VE achieves the highest Recall@5 among all compared models. In contrast, most baseline models rely on frame-level representations (one frame per embedding) with significantly higher embedding dimensionality. This suggests that ML-VE improves recall not by denser representations, but by leveraging coarse aggregation to expand the retrievable candidate set while retaining sufficient semantic discriminability.
\vspace{-0.2cm}
\subsubsection{Evaluation of DiF-SiM}
On the in-house test set, DiF-SiM achieves strong performance across multiple metrics:
\begin{itemize}
    \item F1: 86.79
    \item AP: 94.01
    \item R@P90: 83.32
\end{itemize}
which is achieved not only by DiF-SiM's DeepSim, differential features, and SSL pre-training, but also through the high-quality frame-level embeddings provided by ML-VE.

To extend the comparison to open-source video-deduplication models, we follow the protocol from VCSL \cite{vcsl} and directly use their provided 256-d ISC embeddings as model inputs. Training leverages InternVid \cite{internvid}, Conceptual Captions \cite{cc2m}, and the full VCSL training split, and evaluation is conducted on 55K VCSL test pairs with a balanced distribution of positives and negatives. Compared methods include Hough Voting (\textit{HV}) \cite{hv}, Temporal Networks (\textit{TN}) \cite{tn}, Dynamic Programming (\textit{DP}) \cite{dp}, Dynamic Time Warping (\textit{DTW}) \cite{dtw}, Similarity Pattern Detection (\textit{SPD}) \cite{spd}, TransVCL \cite{transvcl} and Regional Token Representation (\textit{RTR}) \cite{eccv_paper}. 
We adopt the F-score, which is the harmonic mean between the segment-level recall and precision defined by VCSL \cite{vcsl} as the metric to evaluate the effectiveness of temporal localization of copied segments. 

As shown in Table \ref{tab:vcsl-results}, our model achieves an F-score of $74.31\%$, outperforming the previous state of the art (\textit{RTR + pre-training}) by \textbf{3.58}. These results demonstrate that DiF-SiM delivers strong and competitive performance on the public VCSL benchmark, effectively matching and localizing copied video segments under standardized evaluation protocols. 
Visualizations for video copy localization results compared to RTR are shown in Appendix \ref{appendix:video-localization-visualizations}.
\subsection{Ablation Studies}
\subsubsection{Ablations of ML-VE}
We conduct ablation studies on ML-VE by adding a hybrid loss and memory bank deduplication on our baseline backbone, which is a Swin-Large backbone with MoCo-style contrastive loss, regularization loss and inner-feature knowledge distillation. As shown in Table~\ref{tab:ablation_hybrid_memorybank}, introducing a hybrid loss that combines pairwise and triplet objectives improves Recall@5 from 82.46 to 85.72, as it explicitly enforces supervised learning and relative similarity ordering that better matches nearest-neighbor retrieval behavior. Enabling memory bank deduplication helps to mitigate redundant positives and enhance discriminative retrieval under a fixed memory budget, leading to further improvements in Recall@5.
\begin{table}[t]
\centering
\small
\setlength{\tabcolsep}{3pt}
\renewcommand{\arraystretch}{1.03}
\caption{The effectiveness of hybrid loss and memory bank deduplication on ML-VE.}
\label{tab:ablation_hybrid_memorybank}
\begin{tabular}{l |c c c}
\toprule
Recall@5 & Hybrid Loss & Memory Bank Deduplication \\
\midrule
82.46 & \ding{55} &  \ding{55} \\
85.72 & \ding{51} & \ding{55} \\
87.59 & \ding{51} & \ding{51} \\
\bottomrule
\end{tabular}
\end{table}
\begin{table}[t]
\centering
\small
\setlength{\tabcolsep}{2pt}
\renewcommand{\arraystretch}{1.03}
\caption{The effectiveness of inner-feature knowledge distillation (KD) on ML-VE.}
\label{tab:ablation_KD}
\begin{tabular}{l |c c c c c c}
\toprule
Model & AP & F1 & P@R90 & P@R95 & R@P90 & R@P95 \\
\midrule
ML-VE w/o KD  & 91.31 & 83.69 & 66.09 & 47.80 & 77.38 & 74.68 \\
ML-VE & 94.01 & 86.79 & 78.18 & 62.05 & 83.32 & 79.34 \\
Swin-L (teacher) & 94.29 & 87.05 & 79.35 & 63.24 & 84.05 & 78.66 \\
\bottomrule
\end{tabular}
\end{table}
We further evaluate the impact of inner-feature knowledge distillation (KD) in Table~\ref{tab:ablation_KD}. Without KD, ML-VE suffers a clear performance drop in matching-related metrics, especially under strict operating points (i.e., P@R90, P@R95, R@P90, and R@P95). After introducing the KD loss, ML-VE achieves consistent improvements across all metrics, indicating that KD substantially strengthens the quality of frame-level representations used for downstream matching, and narrows the gap to the teacher model.

We provide the loss hyperparameter ablations in Appendix \ref{appendix:loss-hyperparameters}.
\vspace{-0.1cm}
\subsubsection{Similarity threshold in clip-to-video aggregation}
As shown in Table \ref{tab:sim_threshold}, lower thresholds introduce noise, making it harder to rank true duplicates to the top, while higher thresholds miss valid partial matches. We set the threshold to 0.4 for peak performance.
\vspace{-0.1cm}
\begin{table}[t]\small
\centering
\captionsetup{position=top}
\caption{Recall@5 under different similarity thresholds.}
\label{tab:sim_threshold}
\begin{tabular}{lcccccc}
\toprule
\textbf{Similarity threshold} & 0.1 & 0.2 & 0.3 & 0.4 & 0.5 & 0.6 \\
\midrule
\textbf{Recall@5} & 84.52 & 84.92 & 85.32 & \textbf{85.72} & 85.42 & 85.02 \\
\bottomrule
\end{tabular}
\end{table}

\subsubsection{Retrieval Top-K}
As shown in Table \ref{tab:recall_at_k}, recall improves steadily with larger K, from 47.16\% at K=5 to 92.41\% at K=100. The gain diminishes at higher K, indicating most true duplicates are already retrieved at moderate K values.
\begin{table}[t]\small
\centering
\captionsetup{position=top}
\caption{Recall under different $K$ values.}
\label{tab:recall_at_k}
\begin{tabular}{lccccc}
\toprule
\textbf{K} & 5 & 10 & 20 & 50 & 100 \\
\midrule
\textbf{Recall} & 47.16 & 64.12 & 78.19 & 88.75 & 92.41 \\
\bottomrule
\end{tabular}
\end{table}
\vspace{-0.2cm}
\subsubsection{Ablations of DiF-SiM}
\label{section: architectural improvements}
First, we validate the effectiveness of differential features ($\Delta \mathbf{f}_t$), Deep-Sim, and the SSL pre-training strategy on both the open-source VCSL and in-house test sets. For fair comparison, we use 256-d ISC embeddings and train models on VCSL, while 256-d ML-VE frame-level embeddings and corresponding in-house training data are used for business scenarios. As shown in Table~\ref{tab:dif-sim-evolution}, the first row corresponds to the baseline that applies cosine similarity on static frame-level appearance features only. Compared to this baseline, Deep-Sim contributes the dominant performance gain by introducing parametric similarity modeling, which significantly improves all metrics by allowing richer correspondence patterns to be learned beyond a fixed similarity metric. Incorporating $\Delta \mathbf{f}_t$ on top of Deep-Sim further boosts high-precision recall, indicating their effectiveness in capturing temporal transitions and motion consistency that are critical for accurate duplication localization. Image- and video-level self-supervised pre-training progressively stabilize and adapt the learned similarity space to realistic appearance and editing variations. In the experiments, we find that our model can also generalize to unseen distortions (e.g., sticker overlays), as shown in Appendix \ref{sec:similarity_visualization}, but may fail when foreground changes significantly with similar backgrounds.

\begin{table}[t]
\centering
\small
\setlength{\tabcolsep}{2.5pt}
\renewcommand{\arraystretch}{1.03}
\caption{Effectiveness of DiF-SiM components on both the VCSL test set and our in-house test set, ``$\mathbf{f}_t$'', ``$\Delta \mathbf{f}_t$'', ``iPT'', ``vPT''  denote ``static features'', ``differential features'', ``image SSL pre-training'' and ``video SSL pre-training'', respectively.}
\label{tab:dif-sim-evolution}
\begin{tabular}{cccc|ccc|c}
\toprule
\multicolumn{4}{c|}{Method} & \multicolumn{3}{c|}{In-House test set} & \multicolumn{1}{c}{VCSL test set} \\
\hline
\makecell{Deep $\mathbf{f}_t$} &
\makecell{Deep $\Delta \mathbf{f}_t$} &
\makecell{iPT} &
\makecell{vPT} & F1 & AP  & R@P90 & F-score\\
\midrule
\ding{55}     & \ding{55}     & \ding{55}     & \ding{55}     & 83.87 & 91.49 & 77.14 & 70.20 \\
\ding{51} & \ding{55}     & \ding{55}     & \ding{55}     & 84.72 & 92.63 & 79.86 & 72.14 \\
\ding{51} & \ding{51} & \ding{55}     & \ding{55}     & 86.07 & 93.03 & 81.22 & 73.10 \\
\ding{51} & \ding{51} & \ding{51} & \ding{55}     & 86.25 & 93.55 & 81.89 & 73.52 \\
\ding{51} & \ding{51} & \ding{51} & \ding{51} & 86.79 & 94.01 & 83.32 & 74.31 \\
\bottomrule
\end{tabular}
\end{table}
\begin{table}[t]\small
\setlength{\tabcolsep}{2pt}
\caption{Performance after replacing TransVCL's similarity mapping with Deep-Sim on the VCSL test set. ``$\mathbf{f}_t$'' and ``$\Delta \mathbf{f}_t$'' denote static and differential features, respectively.}
\centering
\begin{tabular}{c | c c | ccc}
\toprule
 \makecell{Similarity\\Mapping Method} & \makecell{Similarity\\Map Dim} & Feature & F-score & Precision & Recall  \\
\midrule
\makecell{TransVCL\\Original Attention} & 1  & $\mathbf{f}_t$           & 66.51           &  67.46         & 65.59  \\
Deep-Sim & 8  & $\mathbf{f}_t$           & 70.65           & 68.53          & 72.90  \\
Deep-Sim & 16 & $\mathbf{f}_t$           & 71.59           & 72.75          & 70.48  \\
Deep-Sim & 64 & $\mathbf{f}_t$           & 72.14           & 72.21          & 72.07  \\
Deep-Sim & 8  & $\mathbf{f}_t + \Delta \mathbf{f}_t$    & 72.04           & 69.95          & 74.98  \\
Deep-Sim & 16 & $\mathbf{f}_t + \Delta \mathbf{f}_t$    & 72.62 \         & 70.20          & 75.20  \\ 
Deep-Sim & 64 & $\mathbf{f}_t + \Delta \mathbf{f}_t$    & \textbf{73.10}  & \textbf{73.57} & \textbf{72.62} \\
\bottomrule
\end{tabular}
\label{tab:Deep-SiM_dim_ablation}
\end{table}

Secondly, we validate the generalization ability of Deep-Sim and $\Delta \mathbf{f}_t$. Taking TransVCL as the backbone, we replace the original self-attention and cross-attention modules with Deep-Sim, using either static frame features or $\Delta \mathbf{f}_t$ as input. We illustrate results in Table \ref{tab:Deep-SiM_dim_ablation}. Comparing the second and third row, we find that using only static frame features with Deep-Sim already brings consistent gains over the TransVCL baseline. Incorporating $\Delta \mathbf{f}_t$ further boosts performance across all tested dimensions. At the same dimension, differential versions consistently outperform static ones (i.e., +1.39 F-score at 8d, +1.03 at 16d, +0.96 at 64d). Within each feature type, increasing the frame dimension has a clear positive impact on overall performance. This indicates that higher-dimensional representations allow Deep-Sim to capture more nuanced similarity patterns without overfitting, while the differential input provides complementary information that scales well with dimension. 

On various corruption videos, Deep-Sim and $\Delta \mathbf{f}_t$ also bring gains. We evaluate Recall and F1 without $\Delta \mathbf{f}_t$ or Deep-Sim on 7 augmented test sets (48 pairs each) under common distortions (all perturbations are applied consistently across frames):
\begin{itemize}\small
\item Watermark: random text (5–15 chars) with random size, color, opacity, rotation, and position applied to all frames
\item Random crop: 60–90\% area crop with random position, resized back
\item Compression: 2–5 rounds of JPEG re-encoding (quality 15–40)
\item Temporal scaling (fast): speed-up 1.2–2.5×
\item Temporal scaling (slow): slow-down 0.5–0.9×
\item Gaussian blur: radius 1–9 px
\item Color jitter: brightness/contrast/saturation (0.3–2.0×) and hue shift (-20 to +20)
\end{itemize}
As shown in Table \ref{tab:corruption}, DiF-SiM shows the most stable performance across perturbations compared to w/o $\Delta \mathbf{f}_t$ and w/o both, indicating improved robustness. Deep-Sim provides the main gains (+5.7 F1 avg, up to +12.23 on temporal scaling), showing better robustness than cosine similarity. $\Delta \mathbf{f}_t$ further improves Recall (e.g., +11.96 temporal scaling, +9.13 watermark) by capturing temporal transitions.
\begin{table}[t]\small
\centering
\setlength{\tabcolsep}{4pt}
\caption{Robustness verification of Deep-Sim and $\Delta f_t$ on video corruptions.}
\label{tab:corruption}
\begin{tabular}{lccc}
\toprule
\textbf{\makecell[l]{Transform\\(F1/Recall)}} & \textbf{DiF-SiM} & \textbf{w/o $\Delta f_t$} & \textbf{\makecell{w/o $\Delta f_t$,\\ w/o Deep-Sim}} \\
\midrule
No trans.     & 75.61/84.13 & 74.47/70.90 & 68.77/69.82 \\
Gauss.\ Blur  & 74.19/81.27 & 73.20/72.59 & 66.53/64.93 \\
Watermark     & 75.26/82.10 & 74.71/72.97 & 67.50/67.97 \\
Rand.\ Crop   & 70.24/69.14 & 69.48/68.37 & 65.10/63.29 \\
Color Jitter  & 75.03/80.49 & 74.67/76.32 & 68.06/68.18 \\
Multi.\ Comp. & 75.58/84.11 & 74.39/75.99 & 68.20/68.67 \\
Temp.\ (fast) & 70.88/80.28 & 68.49/68.32 & 56.26/57.25 \\
Temp.\ (slow) & 70.60/72.00 & 69.87/71.18 & 67.81/66.13 \\
\bottomrule
\end{tabular}
\end{table}
We further analyze the influence of YOLO-X module configuration. As shown in Table \ref{tab:width_depth}, increasing width or depth brings only marginal F1 gains but consistently degrades R@P90. (0.33/0.33) achieves the best R@P90 with competitive F1, making it optimal for high-precision deployment.
\vspace{0.1cm}
\begin{table}[t]\small
\centering
\caption{F1 and R@P90 under different YOLO-X width and depth configurations.}
\label{tab:width_depth}
\begin{tabular}{cccc}
\toprule
\textbf{Width} & \textbf{Depth} & \textbf{F1} & \textbf{R@P90} \\
\midrule
0.33 & 0.33 & \textbf{86.07} & \textbf{81.22} \\
0.33 & 1    & 86.35 & 80.87 \\
0.33 & 3    & 85.89 & 80.91 \\
1    & 0.33 & 86.31 & 80.94 \\
3    & 0.33 & 85.62 & 80.53 \\
\bottomrule
\end{tabular}
\end{table}
\section{Related Works}
\subsection{Video Deduplication Pipeline}
Video deduplication pipelines have been extensively studied in multimedia and computer vision \cite{liu2013near, shen2020advance, kordopatis2016near, phalke2018systematic}, driven by applications including copyright protection, web video management, recommendation, and large-scale duplicate filtering. 
With growing video scale, deep learning-based representations and metric learning are applied in the pipeline for better robustness. A prominent recent paradigm adopts a three-stage pipeline: video representation, candidate retrieval, and fine-grained matching \cite{black2023vader, kordopatis2022dns, shen2020advance}.
However, existing pipelines still suffer from limited index capacity in retrieval and imprecise alignment under partial edits and speed changes. To tackle this issue, we adhere to the classic three-stage pipeline while introducing multi-level embeddings for video representation and spatial-temporal matching. Our proposed pipeline expands index capacity with sustained high recall, yielding higher precision and significantly lower duplication rates. 
\vspace{-0.2cm}
\subsection{Video Representation Model}
Recent progress in visual representation learning has substantially improved video feature extraction, driven by strong backbones such as ViT \cite{dosovitskiy2021imageworth16x16words} and Swin Transformer \cite{liu2021swintransformerhierarchicalvision}, as well as self-supervised approaches including MoCo \cite{he2020momentumcontrastunsupervisedvisual,oquab2024dinov2learningrobustvisual}. More recently, large-scale embedding models \cite{muennighoff2025generativerepresentationalinstructiontuning,lee2025nvembedimprovedtechniquestraining,zhang2025gmeimprovinguniversalmultimodal,lin2025mmembeduniversalmultimodalretrieval,jiang2025vlm2vectrainingvisionlanguagemodels,jiang2024e5vuniversalembeddingsmultimodal} further enhance fine-grained frame-level representations. In parallel, multi-frame aggregation architectures such as Video Swin Transformer \cite{liu2021videoswintransformer} and Perceiver \cite{jaegle2021perceivergeneralperceptioniterative} enable compact clip-level representations. While single-frame features achieve strong retrieval performance, they require substantially more indexing capacity, whereas clip-level features are more resource-efficient but tend to yield lower recall and significantly degrade the effectiveness of downstream fine-grained matching modules. Motivated by these advances and real-world deduplication requirements, we design a specialized video representation model that jointly outputs fine-grained frame-level and compact clip-level embeddings. This dual-level representation significantly reduces indexing cost and improves retrieval efficiency, while preserving precise spatial-temporal cues for accurate matching in large-scale online video deduplication.
\vspace{-0.2cm}
\subsection{Matching methods in Deduplication}
Fine-grained matching plays a pivotal role in the video deduplication pipeline, particularly in the third stage of coarse-to-fine architectures, where candidate videos are precisely verified and aligned to identify overlapping segments. Early matching methods (i.e., Hough Voting \cite{hv}, DTW \cite{dtw}, DP \cite{dp}, TN \cite{tn}) relied on fixed formulas (cosine / Euclidean) to build similarity maps, followed by rule-based temporal alignment (voting, path optimization). These methods are rigid and non-adaptive, failing to exploit end-to-end optimization, limiting overall performance in complex, large-scale scenarios.
Recent works (i.e., SPD \cite{spd}, TransVCL \cite{transvcl}, RTR \cite{eccv_paper}) adopt learned similarity modules, achieving better robustness on partial copies.
However, they still use static frame features and ignore differential information, wasting the natural dynamic information in video sequences.
To address this, our method introduces a matching module that explicitly incorporates frame-to-frame differentials into the learned similarity computation for dynamic-aware alignment. By modeling motion residuals as auxiliary tokens, we enhance copy localization under temporal distortions, effectively exploiting video dynamics for more discriminative maps.
\vspace{-0.3cm}
\section{Limitations}
Our method is primarily designed for short-video scenarios, and we acknowledge limitations in challenging cases such as extremely low resolution, heavy blur, or significant foreground changes against highly similar backgrounds. In the future, rather than enumerating distortions through augmentation, we plan to build our representations on large-scale vision foundation models and adapt them with self-supervised contrastive objectives on diverse in-the-wild videos, so that robustness to unseen degradations emerges from broad pretraining rather than hand-crafted augmentations.

In terms of efficiency, compared to Swin-L + cosine matcher on the same A100 hardware, our model increases GPU utilization by ${\sim}23\%$ and VRAM usage by ${\sim}14\%$, mainly due to multi-frame merging in ML-VE's perceiver and the higher dimensionality in DiF-SiM. Nonetheless, the end-to-end pipeline is unchanged and the added overhead is small ($<$200ms), with latency still dominated by network transmission rather than model inference. We plan to reduce this cost via compression techniques such as quantization and distillation, making it negligible even under high-concurrency online serving in the future.
\section{Conclusion}
This paper presents \textbf{MLT-Dedup}, which combines \textbf{M}ulti-\textbf{L}evel video representations with spatial--\textbf{T}emporal matching for large-scale, efficient online video deduplication. We introduce ML-VE to improve candidate coverage under a minimal indexing budget by leveraging task-specific video representations. Two scales of embeddings are generated simultaneously: compact clip-level embeddings for scalable retrieval and frame-level embeddings for fine-grained copy localization. For precise video duplication verification, we introduce \textbf{DiF-SiM}, a \textbf{Di}fferential \textbf{F}eature-enhanced \textbf{Si}milarity \textbf{M}odule, to accurately score recalled pairs. DiF-SiM performs learned similarity modeling and integrates differential features to better capture temporal dynamics within video pairs, and is further strengthened through self-supervised pre-training techniques. Extensive experiments on the public VCSL benchmark and a large in-house dataset show consistent improvements over strong baselines, demonstrating the effectiveness of our system for real-world deduplication. We deploy MLT-Dedup online to serve full production traffic on a large-scale short-video platform and observe significant reductions in video duplication rates.

{
    \small
    \bibliographystyle{ACM-Reference-Format}
    \bibliography{references}


\begin{thebibliography}{46}


\ifx \showCODEN    \undefined \def \showCODEN     #1{\unskip}     \fi
\ifx \showISBNx    \undefined \def \showISBNx     #1{\unskip}     \fi
\ifx \showISBNxiii \undefined \def \showISBNxiii  #1{\unskip}     \fi
\ifx \showISSN     \undefined \def \showISSN      #1{\unskip}     \fi
\ifx \showLCCN     \undefined \def \showLCCN      #1{\unskip}     \fi
\ifx \shownote     \undefined \def \shownote      #1{#1}          \fi
\ifx \showarticletitle \undefined \def \showarticletitle #1{#1}   \fi
\ifx \showURL      \undefined \def \showURL       {\relax}        \fi
\providecommand\bibfield[2]{#2}
\providecommand\bibinfo[2]{#2}
\providecommand\natexlab[1]{#1}
\providecommand\showeprint[2][]{arXiv:#2}

\bibitem[Bardes et~al\mbox{.}(2022)]%
        {bardes2022vicreg}
\bibfield{author}{\bibinfo{person}{Adrien Bardes}, \bibinfo{person}{Jean Ponce}, {and} \bibinfo{person}{Yann LeCun}.} \bibinfo{year}{2022}\natexlab{}.
\newblock \showarticletitle{VICReg: Variance-Invariance-Covariance Regularization For Self-Supervised Learning}. In \bibinfo{booktitle}{\emph{ICLR}}.
\newblock


\bibitem[Berndt and Clifford(1994)]%
        {dtw}
\bibfield{author}{\bibinfo{person}{Donald~J. Berndt} {and} \bibinfo{person}{James Clifford}.} \bibinfo{year}{1994}\natexlab{}.
\newblock \showarticletitle{Using Dynamic Time Warping to Find Patterns in Time Series}. In \bibinfo{booktitle}{\emph{KDD Workshop}}.
\newblock
\urldef\tempurl%
\url{https://api.semanticscholar.org/CorpusID:929893}
\showURL{%
\tempurl}


\bibitem[Black et~al\mbox{.}(2023)]%
        {black2023vader}
\bibfield{author}{\bibinfo{person}{Alexander Black}, \bibinfo{person}{Simon Jenni}, \bibinfo{person}{Tu Bui}, \bibinfo{person}{Md~Mehrab Tanjim}, \bibinfo{person}{Stefano Petrangeli}, \bibinfo{person}{Ritwik Sinha}, \bibinfo{person}{Viswanathan Swaminathan}, {and} \bibinfo{person}{John Collomosse}.} \bibinfo{year}{2023}\natexlab{}.
\newblock \showarticletitle{Vader: Video alignment differencing and retrieval}. In \bibinfo{booktitle}{\emph{Proceedings of the IEEE/CVF International Conference on Computer Vision}}. \bibinfo{pages}{22357--22367}.
\newblock


\bibitem[Chen et~al\mbox{.}(2020)]%
        {chen2020simple}
\bibfield{author}{\bibinfo{person}{Ting Chen}, \bibinfo{person}{Simon Kornblith}, \bibinfo{person}{Mohammad Norouzi}, {and} \bibinfo{person}{Geoffrey Hinton}.} \bibinfo{year}{2020}\natexlab{}.
\newblock \showarticletitle{A simple framework for contrastive learning of visual representations}. In \bibinfo{booktitle}{\emph{International conference on machine learning}}. PmLR, \bibinfo{pages}{1597--1607}.
\newblock


\bibitem[Chou et~al\mbox{.}(2015)]%
        {dp}
\bibfield{author}{\bibinfo{person}{Chien-Li Chou}, \bibinfo{person}{Hua-Tsung Chen}, {and} \bibinfo{person}{Suh-Yin Lee}.} \bibinfo{year}{2015}\natexlab{}.
\newblock \showarticletitle{Pattern-Based Near-Duplicate Video Retrieval and Localization on Web-Scale Videos}.
\newblock \bibinfo{journal}{\emph{IEEE Transactions on Multimedia}} \bibinfo{volume}{17}, \bibinfo{number}{3} (\bibinfo{year}{2015}), \bibinfo{pages}{382--395}.
\newblock
\href{https://doi.org/10.1109/TMM.2015.2391674}{doi:\nolinkurl{10.1109/TMM.2015.2391674}}


\bibitem[de~Sa(1993)]%
        {ssl_first}
\bibfield{author}{\bibinfo{person}{Virginia~R. de Sa}.} \bibinfo{year}{1993}\natexlab{}.
\newblock \showarticletitle{Learning Classification with Unlabeled Data}. In \bibinfo{booktitle}{\emph{Neural Information Processing Systems}}.
\newblock
\urldef\tempurl%
\url{https://api.semanticscholar.org/CorpusID:9890353}
\showURL{%
\tempurl}


\bibitem[Dosovitskiy et~al\mbox{.}(2021)]%
        {dosovitskiy2021imageworth16x16words}
\bibfield{author}{\bibinfo{person}{Alexey Dosovitskiy}, \bibinfo{person}{Lucas Beyer}, \bibinfo{person}{Alexander Kolesnikov}, \bibinfo{person}{Dirk Weissenborn}, \bibinfo{person}{Xiaohua Zhai}, \bibinfo{person}{Thomas Unterthiner}, \bibinfo{person}{Mostafa Dehghani}, \bibinfo{person}{Matthias Minderer}, \bibinfo{person}{Georg Heigold}, \bibinfo{person}{Sylvain Gelly}, \bibinfo{person}{Jakob Uszkoreit}, {and} \bibinfo{person}{Neil Houlsby}.} \bibinfo{year}{2021}\natexlab{}.
\newblock \bibinfo{title}{An Image is Worth 16x16 Words: Transformers for Image Recognition at Scale}.
\newblock
\showeprint[arxiv]{2010.11929}~[cs.CV]
\urldef\tempurl%
\url{https://arxiv.org/abs/2010.11929}
\showURL{%
\tempurl}


\bibitem[Douze et~al\mbox{.}(2010)]%
        {hv}
\bibfield{author}{\bibinfo{person}{Matthijs Douze}, \bibinfo{person}{Herv{\'e} J{\'e}gou}, \bibinfo{person}{Cordelia Schmid}, {and} \bibinfo{person}{Patrick P{\'e}rez}.} \bibinfo{year}{2010}\natexlab{}.
\newblock \showarticletitle{Compact Video Description for Copy Detection with Precise Temporal Alignment}. In \bibinfo{booktitle}{\emph{Computer Vision -- ECCV 2010}}, \bibfield{editor}{\bibinfo{person}{Kostas Daniilidis}, \bibinfo{person}{Petros Maragos}, {and} \bibinfo{person}{Nikos Paragios}} (Eds.). \bibinfo{publisher}{Springer Berlin Heidelberg}, \bibinfo{address}{Berlin, Heidelberg}, \bibinfo{pages}{522--535}.
\newblock
\showISBNx{978-3-642-15549-9}


\bibitem[Fan et~al\mbox{.}(2025)]%
        {fan2025scalinglanguagefreevisualrepresentation}
\bibfield{author}{\bibinfo{person}{David Fan}, \bibinfo{person}{Shengbang Tong}, \bibinfo{person}{Jiachen Zhu}, \bibinfo{person}{Koustuv Sinha}, \bibinfo{person}{Zhuang Liu}, \bibinfo{person}{Xinlei Chen}, \bibinfo{person}{Michael Rabbat}, \bibinfo{person}{Nicolas Ballas}, \bibinfo{person}{Yann LeCun}, \bibinfo{person}{Amir Bar}, {and} \bibinfo{person}{Saining Xie}.} \bibinfo{year}{2025}\natexlab{}.
\newblock \bibinfo{title}{Scaling Language-Free Visual Representation Learning}.
\newblock
\showeprint[arxiv]{2504.01017}~[cs.CV]
\urldef\tempurl%
\url{https://arxiv.org/abs/2504.01017}
\showURL{%
\tempurl}


\bibitem[Ge et~al\mbox{.}(2021)]%
        {yolox}
\bibfield{author}{\bibinfo{person}{Zheng Ge}, \bibinfo{person}{Songtao Liu}, \bibinfo{person}{Feng Wang}, \bibinfo{person}{Zeming Li}, {and} \bibinfo{person}{Jian Sun}.} \bibinfo{year}{2021}\natexlab{}.
\newblock \bibinfo{title}{YOLOX: Exceeding YOLO Series in 2021}.
\newblock
\showeprint[arxiv]{2107.08430}~[cs.CV]
\urldef\tempurl%
\url{https://arxiv.org/abs/2107.08430}
\showURL{%
\tempurl}


\bibitem[Gui et~al\mbox{.}(2024)]%
        {ssl_survey}
\bibfield{author}{\bibinfo{person}{Jie Gui}, \bibinfo{person}{Tuo Chen}, \bibinfo{person}{Jing Zhang}, \bibinfo{person}{Qiong Cao}, \bibinfo{person}{Zhenan Sun}, \bibinfo{person}{Hao Luo}, {and} \bibinfo{person}{Dacheng Tao}.} \bibinfo{year}{2024}\natexlab{}.
\newblock \bibinfo{title}{A Survey on Self-supervised Learning: Algorithms, Applications, and Future Trends}.
\newblock
\showeprint[arxiv]{2301.05712}~[cs.LG]
\urldef\tempurl%
\url{https://arxiv.org/abs/2301.05712}
\showURL{%
\tempurl}


\bibitem[He et~al\mbox{.}(2020)]%
        {he2020momentumcontrastunsupervisedvisual}
\bibfield{author}{\bibinfo{person}{Kaiming He}, \bibinfo{person}{Haoqi Fan}, \bibinfo{person}{Yuxin Wu}, \bibinfo{person}{Saining Xie}, {and} \bibinfo{person}{Ross Girshick}.} \bibinfo{year}{2020}\natexlab{}.
\newblock \bibinfo{title}{Momentum Contrast for Unsupervised Visual Representation Learning}.
\newblock
\showeprint[arxiv]{1911.05722}~[cs.CV]
\urldef\tempurl%
\url{https://arxiv.org/abs/1911.05722}
\showURL{%
\tempurl}


\bibitem[He et~al\mbox{.}(2022a)]%
        {transvcl}
\bibfield{author}{\bibinfo{person}{Sifeng He}, \bibinfo{person}{Yue He}, \bibinfo{person}{Minlong Lu}, \bibinfo{person}{Chen Jiang}, \bibinfo{person}{Xudong Yang}, \bibinfo{person}{Feng Qian}, \bibinfo{person}{Xiaobo Zhang}, \bibinfo{person}{Lei Yang}, {and} \bibinfo{person}{Jiandong Zhang}.} \bibinfo{year}{2022}\natexlab{a}.
\newblock \bibinfo{title}{TransVCL: Attention-enhanced Video Copy Localization Network with Flexible Supervision}.
\newblock
\showeprint[arxiv]{2211.13090}~[cs.CV]
\urldef\tempurl%
\url{https://arxiv.org/abs/2211.13090}
\showURL{%
\tempurl}


\bibitem[He et~al\mbox{.}(2022b)]%
        {vcsl}
\bibfield{author}{\bibinfo{person}{Sifeng He}, \bibinfo{person}{Xudong Yang}, \bibinfo{person}{Chen Jiang}, {et~al\mbox{.}}} \bibinfo{year}{2022}\natexlab{b}.
\newblock \showarticletitle{A Large-scale Comprehensive Dataset and Copy-overlap Aware Evaluation Protocol for Segment-level Video Copy Detection}. In \bibinfo{booktitle}{\emph{Proceedings of the IEEE/CVF Conference on Computer Vision and Pattern Recognition}}. \bibinfo{pages}{21086--21095}.
\newblock


\bibitem[Jaegle et~al\mbox{.}(2021)]%
        {jaegle2021perceivergeneralperceptioniterative}
\bibfield{author}{\bibinfo{person}{Andrew Jaegle}, \bibinfo{person}{Felix Gimeno}, \bibinfo{person}{Andrew Brock}, \bibinfo{person}{Andrew Zisserman}, \bibinfo{person}{Oriol Vinyals}, {and} \bibinfo{person}{Joao Carreira}.} \bibinfo{year}{2021}\natexlab{}.
\newblock \bibinfo{title}{Perceiver: General Perception with Iterative Attention}.
\newblock
\showeprint[arxiv]{2103.03206}~[cs.CV]
\urldef\tempurl%
\url{https://arxiv.org/abs/2103.03206}
\showURL{%
\tempurl}


\bibitem[Jiang et~al\mbox{.}(2021)]%
        {spd}
\bibfield{author}{\bibinfo{person}{Chen Jiang}, \bibinfo{person}{Kaiming Huang}, \bibinfo{person}{Sifeng He}, \bibinfo{person}{Xudong Yang}, \bibinfo{person}{Wei Zhang}, \bibinfo{person}{Xiaobo Zhang}, \bibinfo{person}{Yuan Cheng}, \bibinfo{person}{Lei Yang}, \bibinfo{person}{Qing Wang}, \bibinfo{person}{Furong Xu}, \bibinfo{person}{Tan Pan}, {and} \bibinfo{person}{Wei Chu}.} \bibinfo{year}{2021}\natexlab{}.
\newblock \showarticletitle{Learning Segment Similarity and Alignment in Large-Scale Content Based Video Retrieval}. In \bibinfo{booktitle}{\emph{Proceedings of the 29th ACM International Conference on Multimedia}}. \bibinfo{publisher}{ACM}, \bibinfo{pages}{1618–1626}.
\newblock
\href{https://doi.org/10.1145/3474085.3475301}{doi:\nolinkurl{10.1145/3474085.3475301}}


\bibitem[Jiang et~al\mbox{.}(2024)]%
        {jiang2024e5vuniversalembeddingsmultimodal}
\bibfield{author}{\bibinfo{person}{Ting Jiang}, \bibinfo{person}{Minghui Song}, \bibinfo{person}{Zihan Zhang}, \bibinfo{person}{Haizhen Huang}, \bibinfo{person}{Weiwei Deng}, \bibinfo{person}{Feng Sun}, \bibinfo{person}{Qi Zhang}, \bibinfo{person}{Deqing Wang}, {and} \bibinfo{person}{Fuzhen Zhuang}.} \bibinfo{year}{2024}\natexlab{}.
\newblock \bibinfo{title}{E5-V: Universal Embeddings with Multimodal Large Language Models}.
\newblock
\showeprint[arxiv]{2407.12580}~[cs.CL]
\urldef\tempurl%
\url{https://arxiv.org/abs/2407.12580}
\showURL{%
\tempurl}


\bibitem[Jiang et~al\mbox{.}(2025)]%
        {jiang2025vlm2vectrainingvisionlanguagemodels}
\bibfield{author}{\bibinfo{person}{Ziyan Jiang}, \bibinfo{person}{Rui Meng}, \bibinfo{person}{Xinyi Yang}, \bibinfo{person}{Semih Yavuz}, \bibinfo{person}{Yingbo Zhou}, {and} \bibinfo{person}{Wenhu Chen}.} \bibinfo{year}{2025}\natexlab{}.
\newblock \bibinfo{title}{VLM2Vec: Training Vision-Language Models for Massive Multimodal Embedding Tasks}.
\newblock
\showeprint[arxiv]{2410.05160}~[cs.CV]
\urldef\tempurl%
\url{https://arxiv.org/abs/2410.05160}
\showURL{%
\tempurl}


\bibitem[Johnson et~al\mbox{.}(2017)]%
        {faiss_gpu}
\bibfield{author}{\bibinfo{person}{Jeff Johnson}, \bibinfo{person}{Matthijs Douze}, {and} \bibinfo{person}{Hervé Jégou}.} \bibinfo{year}{2017}\natexlab{}.
\newblock \bibinfo{title}{Billion-scale similarity search with GPUs}.
\newblock
\showeprint[arxiv]{1702.08734}~[cs.CV]
\urldef\tempurl%
\url{https://arxiv.org/abs/1702.08734}
\showURL{%
\tempurl}


\bibitem[Kordopatis-Zilos et~al\mbox{.}(2019)]%
        {kordopatis2019finding}
\bibfield{author}{\bibinfo{person}{Giorgos Kordopatis-Zilos}, \bibinfo{person}{Symeon Papadopoulos}, \bibinfo{person}{Ioannis Patras}, {and} \bibinfo{person}{Ioannis Kompatsiaris}.} \bibinfo{year}{2019}\natexlab{}.
\newblock \showarticletitle{Finding near-duplicate videos in large-scale collections}.
\newblock In \bibinfo{booktitle}{\emph{Video Verification in the Fake News Era}}. \bibinfo{publisher}{Springer}, \bibinfo{pages}{91--126}.
\newblock


\bibitem[Kordopatis-Zilos et~al\mbox{.}(2016)]%
        {kordopatis2016near}
\bibfield{author}{\bibinfo{person}{Giorgos Kordopatis-Zilos}, \bibinfo{person}{Symeon Papadopoulos}, \bibinfo{person}{Ioannis Patras}, {and} \bibinfo{person}{Yiannis Kompatsiaris}.} \bibinfo{year}{2016}\natexlab{}.
\newblock \showarticletitle{Near-duplicate video retrieval by aggregating intermediate cnn layers}. In \bibinfo{booktitle}{\emph{International conference on multimedia modeling}}. Springer, \bibinfo{pages}{251--263}.
\newblock


\bibitem[Kordopatis-Zilos et~al\mbox{.}(2017)]%
        {near_duplicate_video_retreival_with_deep_metric_learning}
\bibfield{author}{\bibinfo{person}{Giorgos Kordopatis-Zilos}, \bibinfo{person}{Symeon Papadopoulos}, \bibinfo{person}{Ioannis Patras}, {and} \bibinfo{person}{Yiannis Kompatsiaris}.} \bibinfo{year}{2017}\natexlab{}.
\newblock \showarticletitle{Near-Duplicate Video Retrieval With Deep Metric Learning}. In \bibinfo{booktitle}{\emph{Proceedings of the IEEE International Conference on Computer Vision (ICCV) Workshops}}.
\newblock


\bibitem[Kordopatis-Zilos et~al\mbox{.}(2022)]%
        {kordopatis2022dns}
\bibfield{author}{\bibinfo{person}{Giorgos Kordopatis-Zilos}, \bibinfo{person}{Christos Tzelepis}, \bibinfo{person}{Symeon Papadopoulos}, \bibinfo{person}{Ioannis Kompatsiaris}, {and} \bibinfo{person}{Ioannis Patras}.} \bibinfo{year}{2022}\natexlab{}.
\newblock \showarticletitle{DnS: Distill-and-select for efficient and accurate video indexing and retrieval}.
\newblock \bibinfo{journal}{\emph{International Journal of Computer Vision}} \bibinfo{volume}{130}, \bibinfo{number}{10} (\bibinfo{year}{2022}), \bibinfo{pages}{2385--2407}.
\newblock


\bibitem[Lee et~al\mbox{.}(2025)]%
        {lee2025nvembedimprovedtechniquestraining}
\bibfield{author}{\bibinfo{person}{Chankyu Lee}, \bibinfo{person}{Rajarshi Roy}, \bibinfo{person}{Mengyao Xu}, \bibinfo{person}{Jonathan Raiman}, \bibinfo{person}{Mohammad Shoeybi}, \bibinfo{person}{Bryan Catanzaro}, {and} \bibinfo{person}{Wei Ping}.} \bibinfo{year}{2025}\natexlab{}.
\newblock \bibinfo{title}{NV-Embed: Improved Techniques for Training LLMs as Generalist Embedding Models}.
\newblock
\showeprint[arxiv]{2405.17428}~[cs.CL]
\urldef\tempurl%
\url{https://arxiv.org/abs/2405.17428}
\showURL{%
\tempurl}


\bibitem[Lin et~al\mbox{.}(2025)]%
        {lin2025mmembeduniversalmultimodalretrieval}
\bibfield{author}{\bibinfo{person}{Sheng-Chieh Lin}, \bibinfo{person}{Chankyu Lee}, \bibinfo{person}{Mohammad Shoeybi}, \bibinfo{person}{Jimmy Lin}, \bibinfo{person}{Bryan Catanzaro}, {and} \bibinfo{person}{Wei Ping}.} \bibinfo{year}{2025}\natexlab{}.
\newblock \bibinfo{title}{MM-Embed: Universal Multimodal Retrieval with Multimodal LLMs}.
\newblock
\showeprint[arxiv]{2411.02571}~[cs.CL]
\urldef\tempurl%
\url{https://arxiv.org/abs/2411.02571}
\showURL{%
\tempurl}


\bibitem[Liu et~al\mbox{.}(2013)]%
        {liu2013near}
\bibfield{author}{\bibinfo{person}{Jiajun Liu}, \bibinfo{person}{Zi Huang}, \bibinfo{person}{Hongyun Cai}, \bibinfo{person}{Heng~Tao Shen}, \bibinfo{person}{Chong~Wah Ngo}, {and} \bibinfo{person}{Wei Wang}.} \bibinfo{year}{2013}\natexlab{}.
\newblock \showarticletitle{Near-duplicate video retrieval: Current research and future trends}.
\newblock \bibinfo{journal}{\emph{ACM Computing Surveys (CSUR)}} \bibinfo{volume}{45}, \bibinfo{number}{4} (\bibinfo{year}{2013}), \bibinfo{pages}{1--23}.
\newblock


\bibitem[Liu et~al\mbox{.}(2015)]%
        {liu2015quantitative}
\bibfield{author}{\bibinfo{person}{Yao Liu}, \bibinfo{person}{Sam Blasiak}, \bibinfo{person}{Weijun Xiao}, \bibinfo{person}{Zhenhua Li}, {and} \bibinfo{person}{Songqing Chen}.} \bibinfo{year}{2015}\natexlab{}.
\newblock \showarticletitle{A Quantitative Study of Video Duplicate Levels in YouTube}. In \bibinfo{booktitle}{\emph{International Conference on Passive and Active Network Measurement}}. Springer, \bibinfo{pages}{235--248}.
\newblock


\bibitem[Liu et~al\mbox{.}(2021a)]%
        {liu2021swintransformerhierarchicalvision}
\bibfield{author}{\bibinfo{person}{Ze Liu}, \bibinfo{person}{Yutong Lin}, \bibinfo{person}{Yue Cao}, \bibinfo{person}{Han Hu}, \bibinfo{person}{Yixuan Wei}, \bibinfo{person}{Zheng Zhang}, \bibinfo{person}{Stephen Lin}, {and} \bibinfo{person}{Baining Guo}.} \bibinfo{year}{2021}\natexlab{a}.
\newblock \bibinfo{title}{Swin Transformer: Hierarchical Vision Transformer using Shifted Windows}.
\newblock
\showeprint[arxiv]{2103.14030}~[cs.CV]
\urldef\tempurl%
\url{https://arxiv.org/abs/2103.14030}
\showURL{%
\tempurl}


\bibitem[Liu et~al\mbox{.}(2021b)]%
        {liu2021videoswintransformer}
\bibfield{author}{\bibinfo{person}{Ze Liu}, \bibinfo{person}{Jia Ning}, \bibinfo{person}{Yue Cao}, \bibinfo{person}{Yixuan Wei}, \bibinfo{person}{Zheng Zhang}, \bibinfo{person}{Stephen Lin}, {and} \bibinfo{person}{Han Hu}.} \bibinfo{year}{2021}\natexlab{b}.
\newblock \bibinfo{title}{Video Swin Transformer}.
\newblock
\showeprint[arxiv]{2106.13230}~[cs.CV]
\urldef\tempurl%
\url{https://arxiv.org/abs/2106.13230}
\showURL{%
\tempurl}


\bibitem[Loshchilov and Hutter(2019)]%
        {adamW}
\bibfield{author}{\bibinfo{person}{Ilya Loshchilov} {and} \bibinfo{person}{Frank Hutter}.} \bibinfo{year}{2019}\natexlab{}.
\newblock \bibinfo{title}{Decoupled Weight Decay Regularization}.
\newblock
\showeprint[arxiv]{1711.05101}~[cs.LG]
\urldef\tempurl%
\url{https://arxiv.org/abs/1711.05101}
\showURL{%
\tempurl}


\bibitem[Lu et~al\mbox{.}(2025)]%
        {eccv_paper}
\bibfield{author}{\bibinfo{person}{Minlong Lu}, \bibinfo{person}{Yichen Lu}, \bibinfo{person}{Siwei Nie}, \bibinfo{person}{Xudong Yang}, {and} \bibinfo{person}{Xiaobo Zhang}.} \bibinfo{year}{2025}\natexlab{}.
\newblock \showarticletitle{Self-supervised Video Copy Localization with Regional Token Representation}. In \bibinfo{booktitle}{\emph{Computer Vision -- ECCV 2024}}, \bibfield{editor}{\bibinfo{person}{Ale{\v{s}} Leonardis}, \bibinfo{person}{Elisa Ricci}, \bibinfo{person}{Stefan Roth}, \bibinfo{person}{Olga Russakovsky}, \bibinfo{person}{Torsten Sattler}, {and} \bibinfo{person}{G{\"u}l Varol}} (Eds.). \bibinfo{publisher}{Springer Nature Switzerland}, \bibinfo{address}{Cham}, \bibinfo{pages}{18--35}.
\newblock
\showISBNx{978-3-031-73254-6}


\bibitem[Malkov and Yashunin(2018)]%
        {hnsw}
\bibfield{author}{\bibinfo{person}{Yu.~A. Malkov} {and} \bibinfo{person}{D.~A. Yashunin}.} \bibinfo{year}{2018}\natexlab{}.
\newblock \bibinfo{title}{Efficient and robust approximate nearest neighbor search using Hierarchical Navigable Small World graphs}.
\newblock
\showeprint[arxiv]{1603.09320}~[cs.DS]
\urldef\tempurl%
\url{https://arxiv.org/abs/1603.09320}
\showURL{%
\tempurl}


\bibitem[Muennighoff et~al\mbox{.}(2025)]%
        {muennighoff2025generativerepresentationalinstructiontuning}
\bibfield{author}{\bibinfo{person}{Niklas Muennighoff}, \bibinfo{person}{Hongjin Su}, \bibinfo{person}{Liang Wang}, \bibinfo{person}{Nan Yang}, \bibinfo{person}{Furu Wei}, \bibinfo{person}{Tao Yu}, \bibinfo{person}{Amanpreet Singh}, {and} \bibinfo{person}{Douwe Kiela}.} \bibinfo{year}{2025}\natexlab{}.
\newblock \bibinfo{title}{Generative Representational Instruction Tuning}.
\newblock
\showeprint[arxiv]{2402.09906}~[cs.CL]
\urldef\tempurl%
\url{https://arxiv.org/abs/2402.09906}
\showURL{%
\tempurl}


\bibitem[Oquab et~al\mbox{.}(2024)]%
        {oquab2024dinov2learningrobustvisual}
\bibfield{author}{\bibinfo{person}{Maxime Oquab}, \bibinfo{person}{Timothée Darcet}, \bibinfo{person}{Théo Moutakanni}, \bibinfo{person}{Huy Vo}, \bibinfo{person}{Marc Szafraniec}, \bibinfo{person}{Vasil Khalidov}, \bibinfo{person}{Pierre Fernandez}, \bibinfo{person}{Daniel Haziza}, \bibinfo{person}{Francisco Massa}, \bibinfo{person}{Alaaeldin El-Nouby}, \bibinfo{person}{Mahmoud Assran}, \bibinfo{person}{Nicolas Ballas}, \bibinfo{person}{Wojciech Galuba}, \bibinfo{person}{Russell Howes}, \bibinfo{person}{Po-Yao Huang}, \bibinfo{person}{Shang-Wen Li}, \bibinfo{person}{Ishan Misra}, \bibinfo{person}{Michael Rabbat}, \bibinfo{person}{Vasu Sharma}, \bibinfo{person}{Gabriel Synnaeve}, \bibinfo{person}{Hu Xu}, \bibinfo{person}{Hervé Jegou}, \bibinfo{person}{Julien Mairal}, \bibinfo{person}{Patrick Labatut}, \bibinfo{person}{Armand Joulin}, {and} \bibinfo{person}{Piotr Bojanowski}.} \bibinfo{year}{2024}\natexlab{}.
\newblock \bibinfo{title}{DINOv2: Learning Robust Visual Features without Supervision}.
\newblock
\showeprint[arxiv]{2304.07193}~[cs.CV]
\urldef\tempurl%
\url{https://arxiv.org/abs/2304.07193}
\showURL{%
\tempurl}


\bibitem[Phalke et~al\mbox{.}(2018)]%
        {phalke2018systematic}
\bibfield{author}{\bibinfo{person}{Dhanashree~A Phalke}, \bibinfo{person}{Sunita Jahirabadkar}, {and} \bibinfo{person}{SPPU Pune}.} \bibinfo{year}{2018}\natexlab{}.
\newblock \showarticletitle{A systematic review of near duplicate video retrieval techniques}.
\newblock \bibinfo{journal}{\emph{International Journal of Pure and Applied Mathematics}} \bibinfo{volume}{118}, \bibinfo{number}{24} (\bibinfo{year}{2018}), \bibinfo{pages}{1--11}.
\newblock


\bibitem[Rodrigues et~al\mbox{.}(2010)]%
        {rodrigues2010equal}
\bibfield{author}{\bibinfo{person}{Tiago Rodrigues}, \bibinfo{person}{Fabr{\'\i}cio Benevenuto}, \bibinfo{person}{Virg{\'\i}lio Almeida}, \bibinfo{person}{Jussara Almeida}, {and} \bibinfo{person}{Marcos Gon{\c{c}}alves}.} \bibinfo{year}{2010}\natexlab{}.
\newblock \showarticletitle{Equal but different: a contextual analysis of duplicated videos on YouTube}.
\newblock \bibinfo{journal}{\emph{Journal of the Brazilian Computer Society}} \bibinfo{volume}{16}, \bibinfo{number}{3} (\bibinfo{year}{2010}), \bibinfo{pages}{201--214}.
\newblock


\bibitem[Schuhmann et~al\mbox{.}(2022)]%
        {schuhmann2022laion}
\bibfield{author}{\bibinfo{person}{Christoph Schuhmann}, \bibinfo{person}{Romain Beaumont}, \bibinfo{person}{Richard Vencu}, \bibinfo{person}{Cade Gordon}, \bibinfo{person}{Ross Wightman}, \bibinfo{person}{Mehdi Cherti}, \bibinfo{person}{Theo Coombes}, \bibinfo{person}{Aarush Katta}, \bibinfo{person}{Clayton Mullis}, \bibinfo{person}{Mitchell Wortsman}, {et~al\mbox{.}}} \bibinfo{year}{2022}\natexlab{}.
\newblock \showarticletitle{Laion-5b: An open large-scale dataset for training next generation image-text models}.
\newblock \bibinfo{journal}{\emph{Advances in neural information processing systems}}  \bibinfo{volume}{35} (\bibinfo{year}{2022}), \bibinfo{pages}{25278--25294}.
\newblock


\bibitem[Sharma et~al\mbox{.}(2018)]%
        {cc2m}
\bibfield{author}{\bibinfo{person}{Piyush Sharma}, \bibinfo{person}{Nan Ding}, \bibinfo{person}{Sebastian Goodman}, {and} \bibinfo{person}{Radu Soricut}.} \bibinfo{year}{2018}\natexlab{}.
\newblock \showarticletitle{Conceptual Captions: A Cleaned, Hypernymed, Image Alt-text Dataset For Automatic Image Captioning}. In \bibinfo{booktitle}{\emph{Proceedings of ACL}}.
\newblock


\bibitem[Shen et~al\mbox{.}(2020)]%
        {shen2020advance}
\bibfield{author}{\bibinfo{person}{Ling Shen}, \bibinfo{person}{Richang Hong}, {and} \bibinfo{person}{Yanbin Hao}.} \bibinfo{year}{2020}\natexlab{}.
\newblock \showarticletitle{Advance on large scale near-duplicate video retrieval}.
\newblock \bibinfo{journal}{\emph{Frontiers of Computer Science}} \bibinfo{volume}{14}, \bibinfo{number}{5} (\bibinfo{year}{2020}), \bibinfo{pages}{145702}.
\newblock


\bibitem[Tan et~al\mbox{.}(2009)]%
        {tn}
\bibfield{author}{\bibinfo{person}{Hung-Khoon Tan}, \bibinfo{person}{Chong-Wah Ngo}, \bibinfo{person}{Richard Hong}, {and} \bibinfo{person}{Tat-Seng Chua}.} \bibinfo{year}{2009}\natexlab{}.
\newblock \showarticletitle{Scalable detection of partial near-duplicate videos by visual-temporal consistency}. In \bibinfo{booktitle}{\emph{Proceedings of the 17th ACM International Conference on Multimedia}} (Beijing, China) \emph{(\bibinfo{series}{MM '09})}. \bibinfo{publisher}{Association for Computing Machinery}, \bibinfo{address}{New York, NY, USA}, \bibinfo{pages}{145–154}.
\newblock
\showISBNx{9781605586083}
\href{https://doi.org/10.1145/1631272.1631295}{doi:\nolinkurl{10.1145/1631272.1631295}}


\bibitem[Wang et~al\mbox{.}(2024)]%
        {internvid}
\bibfield{author}{\bibinfo{person}{Yi Wang}, \bibinfo{person}{Yinan He}, \bibinfo{person}{Yizhuo Li}, \bibinfo{person}{Kunchang Li}, \bibinfo{person}{Jiashuo Yu}, \bibinfo{person}{Xin Ma}, \bibinfo{person}{Xinhao Li}, \bibinfo{person}{Guo Chen}, \bibinfo{person}{Xinyuan Chen}, \bibinfo{person}{Yaohui Wang}, \bibinfo{person}{Conghui He}, \bibinfo{person}{Ping Luo}, \bibinfo{person}{Ziwei Liu}, \bibinfo{person}{Yali Wang}, \bibinfo{person}{Limin Wang}, {and} \bibinfo{person}{Yu Qiao}.} \bibinfo{year}{2024}\natexlab{}.
\newblock \bibinfo{title}{InternVid: A Large-scale Video-Text Dataset for Multimodal Understanding and Generation}.
\newblock
\showeprint[arxiv]{2307.06942}~[cs.CV]
\urldef\tempurl%
\url{https://arxiv.org/abs/2307.06942}
\showURL{%
\tempurl}


\bibitem[Wu et~al\mbox{.}(2007)]%
        {practical_elimination_of_near_duplicates_from_web_video_search}
\bibfield{author}{\bibinfo{person}{Xiao Wu}, \bibinfo{person}{Alexander~G. Hauptmann}, {and} \bibinfo{person}{Chong-Wah Ngo}.} \bibinfo{year}{2007}\natexlab{}.
\newblock \showarticletitle{Practical elimination of near-duplicates from web video search}. In \bibinfo{booktitle}{\emph{Proceedings of the 15th ACM International Conference on Multimedia}} (Augsburg, Germany) \emph{(\bibinfo{series}{MM '07})}. \bibinfo{publisher}{Association for Computing Machinery}, \bibinfo{address}{New York, NY, USA}, \bibinfo{pages}{218–227}.
\newblock
\showISBNx{9781595937025}
\href{https://doi.org/10.1145/1291233.1291280}{doi:\nolinkurl{10.1145/1291233.1291280}}


\bibitem[Yew et~al\mbox{.}(2025)]%
        {yew2025dynamic}
\bibfield{author}{\bibinfo{person}{Wei~Chee Yew}, \bibinfo{person}{Hailun Xu}, \bibinfo{person}{Sanjay Saha}, \bibinfo{person}{Xiaotian Fan}, \bibinfo{person}{Hiok~Hian Ong}, \bibinfo{person}{David~Yuchen Wang}, \bibinfo{person}{Kanchan Sarkar}, \bibinfo{person}{Zhenheng Yang}, {and} \bibinfo{person}{Danhui Guan}.} \bibinfo{year}{2025}\natexlab{}.
\newblock \showarticletitle{Dynamic Content Moderation in Livestreams: Combining Supervised Classification with MLLM-Boosted Similarity Matching}.
\newblock \bibinfo{journal}{\emph{arXiv preprint arXiv:2512.03553}} (\bibinfo{year}{2025}).
\newblock


\bibitem[Zhang et~al\mbox{.}(2025)]%
        {zhang2025gmeimprovinguniversalmultimodal}
\bibfield{author}{\bibinfo{person}{Xin Zhang}, \bibinfo{person}{Yanzhao Zhang}, \bibinfo{person}{Wen Xie}, \bibinfo{person}{Mingxin Li}, \bibinfo{person}{Ziqi Dai}, \bibinfo{person}{Dingkun Long}, \bibinfo{person}{Pengjun Xie}, \bibinfo{person}{Meishan Zhang}, \bibinfo{person}{Wenjie Li}, {and} \bibinfo{person}{Min Zhang}.} \bibinfo{year}{2025}\natexlab{}.
\newblock \bibinfo{title}{GME: Improving Universal Multimodal Retrieval by Multimodal LLMs}.
\newblock
\showeprint[arxiv]{2412.16855}~[cs.CL]
\urldef\tempurl%
\url{https://arxiv.org/abs/2412.16855}
\showURL{%
\tempurl}


\bibitem[Zhu et~al\mbox{.}(2026a)]%
        {zhu2026camel}
\bibfield{author}{\bibinfo{person}{Zirui Zhu}, \bibinfo{person}{Hailun Xu}, \bibinfo{person}{Yang Luo}, \bibinfo{person}{Yong Liu}, \bibinfo{person}{Kanchan Sarkar}, \bibinfo{person}{Kun Xu}, {and} \bibinfo{person}{Yang You}.} \bibinfo{year}{2026}\natexlab{a}.
\newblock \showarticletitle{CAMEL: Confidence-Gated Reflection for Reward Modeling}.
\newblock \bibinfo{journal}{\emph{arXiv preprint arXiv:2602.20670}} (\bibinfo{year}{2026}).
\newblock


\bibitem[Zhu et~al\mbox{.}(2026b)]%
        {zhu2026focus}
\bibfield{author}{\bibinfo{person}{Zirui Zhu}, \bibinfo{person}{Hailun Xu}, \bibinfo{person}{Yang Luo}, \bibinfo{person}{Yong Liu}, \bibinfo{person}{Kanchan Sarkar}, \bibinfo{person}{Zhenheng Yang}, {and} \bibinfo{person}{Yang You}.} \bibinfo{year}{2026}\natexlab{b}.
\newblock \showarticletitle{FOCUS: Efficient Keyframe Selection for Long Video Understanding}. In \bibinfo{booktitle}{\emph{International Conference on Learning Representations}}.
\newblock


\end{thebibliography}
}
\onecolumn
\newpage
\appendix
\setcounter{figure}{0}
\renewcommand{\thefigure}{A\arabic{figure}}
\setcounter{table}{0}
\renewcommand{\thetable}{A\arabic{table}}
\section*{Appendix}
\addcontentsline{toc}{section}{Appendix}
\section{Differential Features vs. Static Features}
\label{sec:similarity_visualization}
\begin{figure}[h!]
\centering
\includegraphics[width=0.8\textwidth,trim={0cm 5cm 0cm 2cm}, clip]{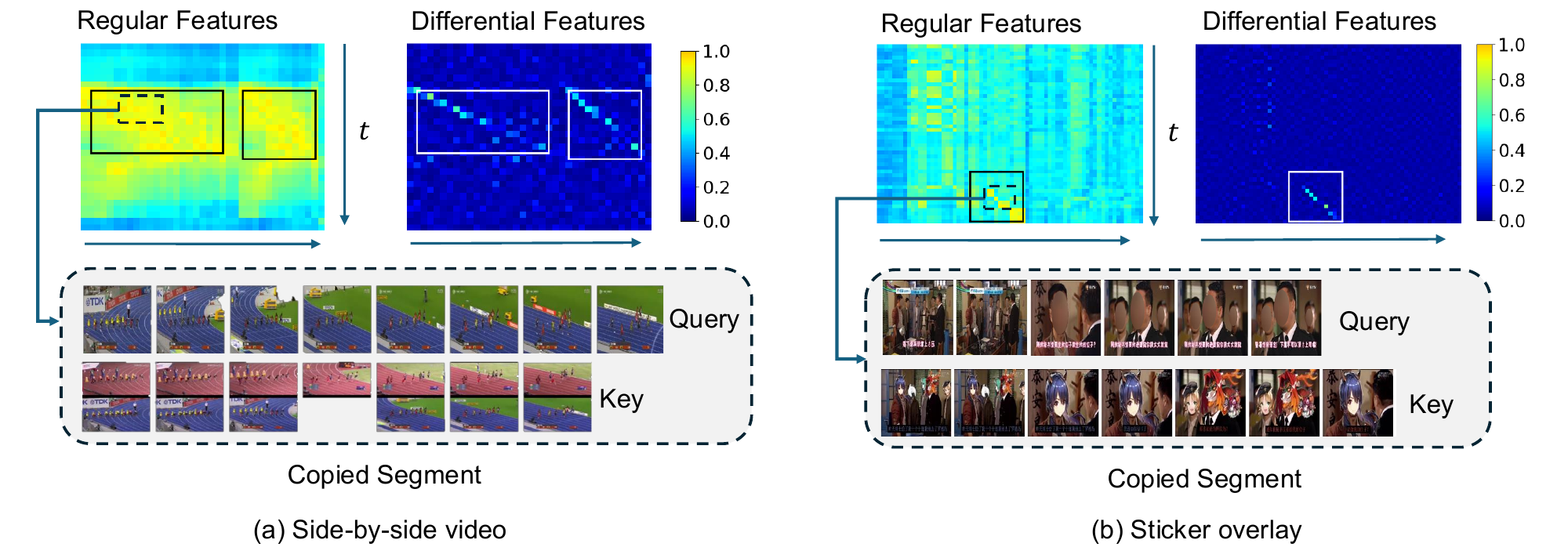}
\caption{Similarity map from differential features compared to static features.} 
\Description{Two visual examples comparing regular-feature similarity maps with differential-feature similarity maps under side-by-side composition and sticker overlay distortions.}
\label{fig:diff_vs_static}
\end{figure}

We visualize the normalized similarity map formed from regular features and differential features. Figure \ref{fig:diff_vs_static} presents two examples of copied video pairs subjected to different types of distortion: (a) a side-by-side video composition and (b) a sticker overlay. In both cases, differential features exhibit a significantly cleaner pattern, highlighting potential regions of video duplication far more effectively than the regular features. Notably, the sticker overlay in (b) is a distortion type that is \emph{not} included in our augmentation strategy; nevertheless, the differential-feature formulation is still able to detect the copied segment, demonstrating its generalization to unseen distortion types.

\section{DiF-SiM Self-Supervised Pre-training with Dual-Modality Data Augmentation}
\label{appendix:ssl-aug}
\begin{figure*}[h]
    \centering
    \begin{subfigure}[t]{0.25\textwidth}
        \centering
        \includegraphics[width=\textwidth]{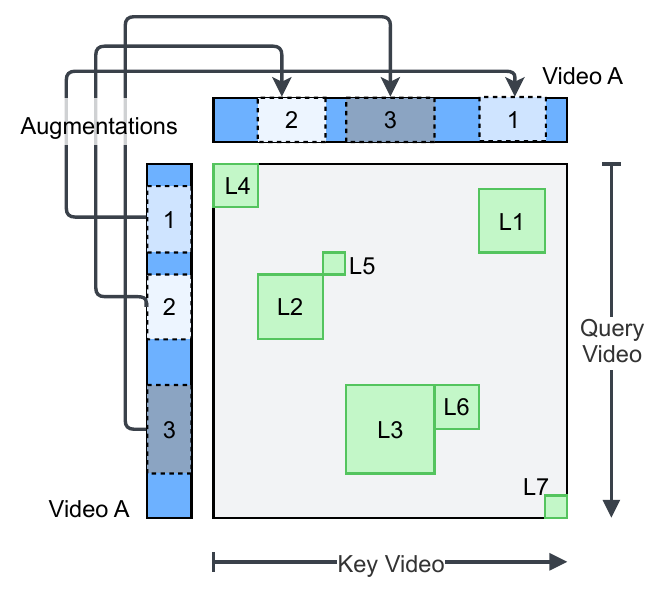}
        \captionsetup{width=0.8\textwidth}
        \caption{Self-insertion.}
        \label{AA-transform}
    \end{subfigure}
    \hfill
    \begin{subfigure}[t]{0.35\textwidth}
        \centering
        \includegraphics[width=\textwidth]{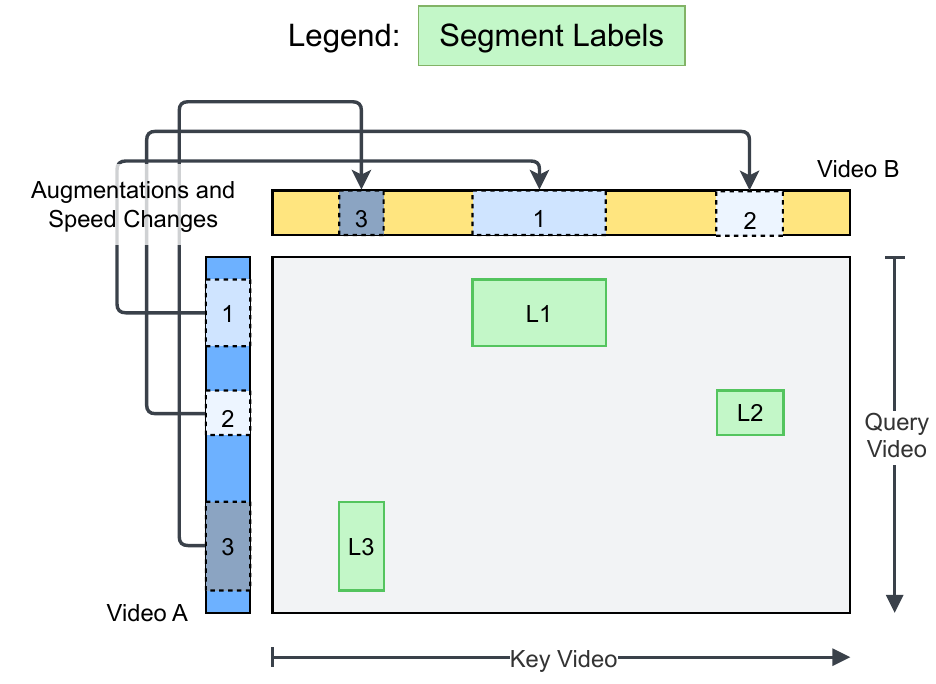}
        \captionsetup{width=0.8\textwidth}
        \caption{Single-insertion.}
     \label{AB-transform}
    \end{subfigure}
    \hfill
    \begin{subfigure}[t]{0.39\textwidth}
        \centering
        \includegraphics[width=\textwidth]{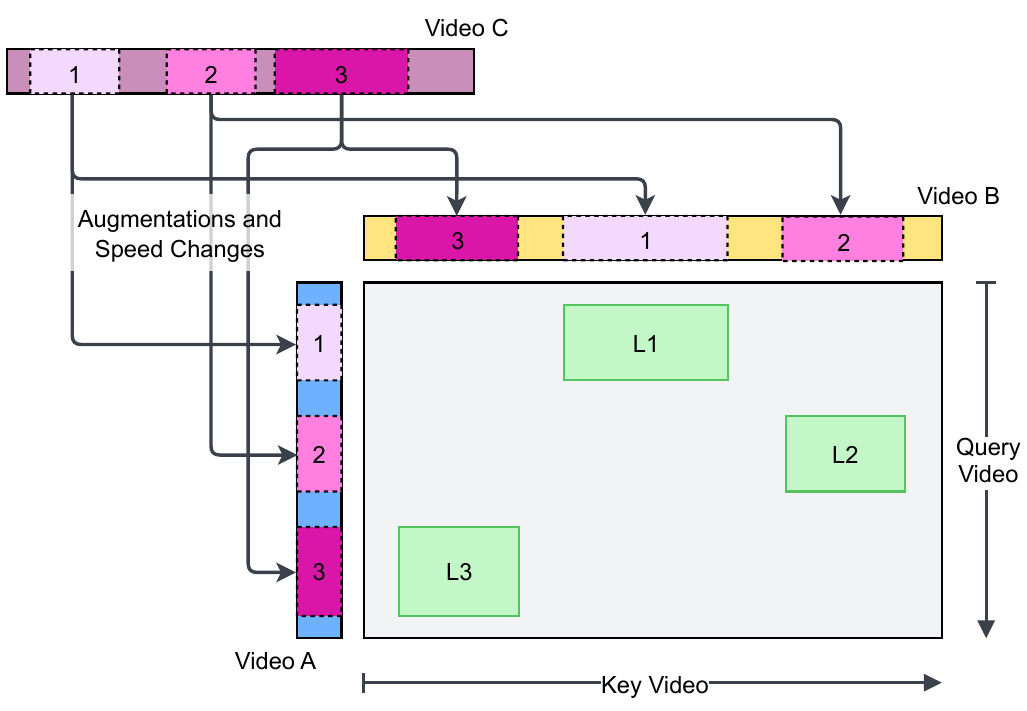}
        \captionsetup{width=0.8\textwidth}
        \caption{Double-insertion.}

     \label{ABC-transform}
    \end{subfigure}
    \caption{Illustrations of the augmentations utilized for self-supervised training. (a) Self-insertion. Both the query and key videos are created from a single video A. Random segments from the query are replaced into the key, creating segment labels L1-L3. Since query and key are the same video, the remaining un-replaced key segments also form copied segments, shown as L4-L7. (b) Single-insertion. The query and key videos are created from two videos A and B. Random segments from the query are inserted into the key to form segment labels L1-L3. (c) Double-insertion. The query and key are derived from two videos A and B. Random segments from a third video C are inserted into both the query and key to form segment labels L1-L3.}
    \Description{Three augmentation diagrams showing self-insertion, single-insertion, and double-insertion transformations used for self-supervised video pre-training.}
    \label{fig:ssl-transforms}
\end{figure*}

We visualize the self-supervised augmentations used in our video pre-training for DiF-SiM. Figures \ref{AA-transform}, \ref{AB-transform}, and \ref{ABC-transform} show self-insertion transform, single-insertion transform, and double-insertion transform, respectively. 

\section{Video Localization Result Visualizations} 
\label{appendix:video-localization-visualizations}

\begin{figure}[h!]
\centering
\begin{subfigure}[b]{0.48\textwidth}
    \centering
    \includegraphics[width=\textwidth]{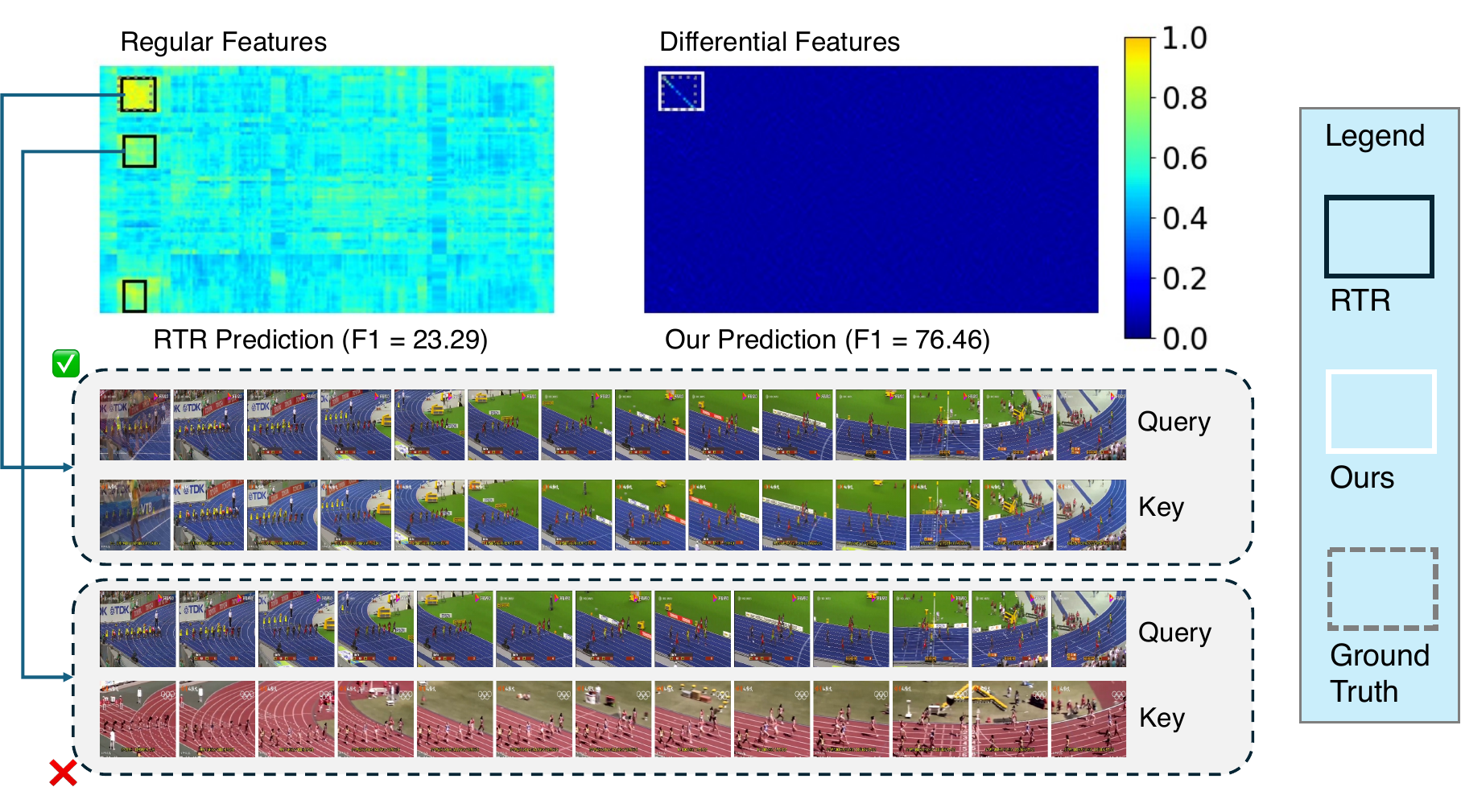}
    \caption{}
    \label{fig:Localization1}
\end{subfigure}
\hfill
\begin{subfigure}[b]{0.48\textwidth}
    \centering
    \includegraphics[width=\textwidth]{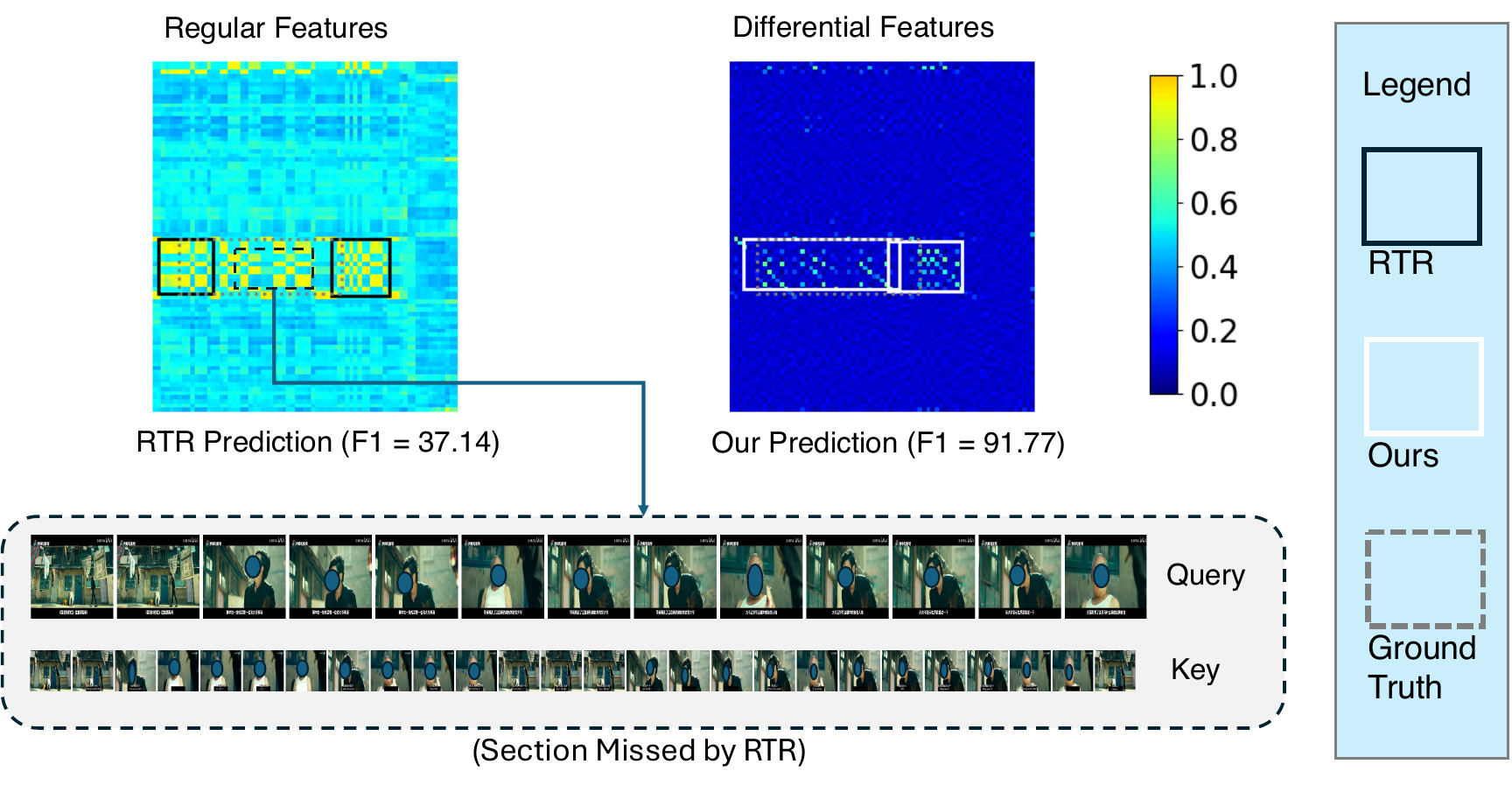}
    \caption{}
    \label{fig:Localization2}
\end{subfigure}
\caption{Two examples of video copy localization predictions made by the RTR model and our model. In each subfigure, the cosine similarity map is visualized on the left and the differential map is visualized on the right. In (a), we show the video frames for the correctly predicted section (top) and the incorrectly predicted section by RTR (bottom). We see that both of these sections exhibit high similarity, but the latter pair is not a duplicate pair. In (b), the duplicated video section is missed by RTR. Our differential features and DiF-SiM module help to more clearly differentiate between such hard cases to make more accurate predictions, achieving a much higher F1 score than RTR.}
\Description{Two localization examples comparing RTR predictions with the proposed model using cosine similarity maps, differential maps, and selected video frames.}
\label{case-1}
\end{figure}

We show visualization examples on the VCSL video localization test set using our proposed DiF-SiM matching model. We visualize and compare them with predictions made by RTR \cite{eccv_paper}. The visualizations shown in Figure \ref{case-1} demonstrate that our differential features and DiF-SiM effectively resolve ambiguous high-similarity cases where RTR fails, leading to more accurate temporal copy localization and a substantially higher F1 score.

\FloatBarrier

\section{Ablations of Loss Hyperparameters} 
\label{appendix:loss-hyperparameters}
Losses except KD affect retrieval (Recall@5) with little impact on matching F1, so we report Recall@5 in the following table. MoCo is the base objective. VICReg improves retrieval at small weights ($\alpha=0.1$) by preventing collapse, while larger weights give limited gains. Triplet loss contributes more than pairwise as it directly optimizes ranking; combining both further improves performance. KD mainly affects F1 rather than Recall@5 because it targets frame-level feature matching. We report the influence of each hyperparameter in Table \ref{tab:weight_ablation}.

\begin{table}[htbp]\small
\centering
\begin{subtable}[t]{0.62\textwidth}
    \centering
    \resizebox{\textwidth}{!}{
    \begin{tabular}{cccccc}
    \toprule
    \textbf{MoCo} & \textbf{VICReg ($\alpha$)} & \textbf{Pairwise ($\beta$)} & \textbf{Triplet ($\rho$)} & \textbf{KD ($\omega$)} & \textbf{Recall@5} \\
    \midrule
    1 & 0   & 0   & 0   & 0   & 80.49 \\
    1 & 0.1 & 0   & 0   & 0   & 82.43 \\
    1 & 1   & 0   & 0   & 0   & 82.15 \\
    1 & 10  & 0   & 0   & 0   & 81.76 \\
    1 & 0   & 1   & 0   & 0   & 82.38 \\
    1 & 0   & 0   & 1   & 0   & 83.07 \\
    1 & 0   & 0.1 & 0.1 & 0   & 84.19 \\
    1 & 0   & 1   & 1   & 0   & 84.56 \\
    1 & 0   & 10  & 10  & 0   & 83.27 \\
    1 & 0   & 0   & 0   & 0.1 & 80.64 \\
    1 & 0   & 0   & 0   & 1   & 80.57 \\
    1 & 0   & 0   & 0   & 10  & 78.91 \\
    1 & 0.1 & 1   & 1   & 1   & \textbf{85.72} \\
    \bottomrule
    \end{tabular}}
    \caption{Recall@5 under different loss weight combinations.}
    \label{tab:loss_weights}
\end{subtable}
\hfill
\begin{subtable}[t]{0.35\textwidth}
    \centering
    \begin{tabular}{ccc}
    \toprule
    \textbf{MoCo} & \textbf{KD ($\omega$)} & \textbf{F1} \\
    \midrule
    1 & 0.1 & 85.86 \\
    1 & 1   & 86.79 \\
    1 & 10  & \textbf{87.65} \\
    \bottomrule
    \end{tabular}
    \caption{The impact of KD weight $\omega$ on F1.}
    \label{tab:kd_weight}
\end{subtable}
\caption{Ablation study of loss hyperparameters.}
\label{tab:weight_ablation}
\end{table}

\end{document}